\documentclass[10pt,twocolumn,letterpaper]{article}

\usepackage{cvpr}              % To produce the CAMERA-READY version
\usepackage{amsmath, amssymb, amsthm}
\usepackage{booktabs}
\usepackage{siunitx}
\usepackage{graphicx}
\usepackage{xcolor}
\usepackage{multicol}
\usepackage{colortbl}
\usepackage{tabularx}
\usepackage{makecell}
\usepackage{multirow}
\usepackage{diagbox}
\usepackage{wrapfig}
\usepackage{enumitem}
\usepackage{algorithm}
\usepackage{algpseudocode}
\usepackage[normalem]{ulem}
\usepackage{tgcursor}
\usepackage{bbm}
\usepackage[utf8]{inputenc}
\usepackage{amsmath}
\usepackage{amssymb}

\newcommand{\mymethod}{SDC}

\algrenewcommand{\algorithmicrequire}{\textbf{Input:}}
\algrenewcommand{\algorithmicensure}{\textbf{Output:}}

\newcommand{\easy}[1]{\texttt{easy}}
\newcommand{\hard}[1]{\texttt{hard}}

\definecolor{cvprblue}{rgb}{0.21,0.49,0.74}
\usepackage[pagebackref,breaklinks,colorlinks,allcolors=cvprblue]{hyperref}

\def\paperID{13} % *** Enter the Paper ID here
\def\confName{CVPR}
\def\confYear{2025}

\title{Not All Samples Should Be Utilized Equally: Towards Understanding and Improving Dataset Distillation}

\author{
    Shaobo Wang$^{1,2}$ \quad Yantai Yang$^{2}$ \quad  Qilong Wang$^2$  \quad Kaixin Li$^3$ \\
     \quad Linfeng Zhang$^{1,2}$ \quad Junchi Yan$^1$\thanks{Corresponding Author.}
    \\ $^1$School of Artificial Intelligence, Shanghai Jiao Tong University \\
      $^2$EPIC Lab, Shanghai Jiao Tong University   \\
      $^3$National University of Singapore \\
      \\  \texttt{\{shaobowang1009,yanjunchi\}@sjtu.edu.cn}
}

\begin{document}
\maketitle

\newtheorem{theorem}{Theorem}
\newtheorem{lemma}{Lemma}
\newtheorem{observation}{Observation}
\newtheorem{analysis}{Analysis}
\newtheorem{definition}{Definition}
\newtheorem{corollary}{Corollary}
\newtheorem{proposition}{Proposition}

\begin{abstract}
\textit{Dataset Distillation} (DD) aims to synthesize a small dataset capable of performing comparably to the original dataset. Despite the success of numerous DD methods, theoretical exploration of this area remains unaddressed. In this paper, we take an initial step towards understanding various matching-based DD methods from the perspective of \textit{sample difficulty}. We begin by empirically examining sample difficulty, measured by gradient norm, and observe that different matching-based methods roughly correspond to specific difficulty tendencies. We then extend the neural scaling laws of data pruning to DD to theoretically explain these matching-based methods. Our findings suggest that prioritizing the synthesis of easier samples from the original dataset can enhance the quality of distilled datasets, especially in low IPC (image-per-class) settings. Based on our empirical observations and theoretical analysis, we introduce the \textit{Sample Difficulty Correction} (SDC) approach, designed to predominantly generate easier samples to achieve higher dataset quality. Our SDC can be seamlessly integrated into existing methods as a plugin with minimal code adjustments. Experimental results demonstrate that adding SDC  generates higher-quality distilled datasets across 7 distillation methods and 6 datasets. 
\end{abstract}    
\section{Introduction}
In an era of data-centric AI, scaling laws \cite{kaplan2020scaling} have shifted the focus to data quality. Under this scenario, dataset distillation (DD) \cite{wang2018dataset, sachdeva2023data,DRUPI,NCFM} has emerged as a solution for creating high-quality data summaries. Unlike data pruning methods \cite{ghorbani2019data, coleman2019selection, toneva2018empirical, abbas2023semdedup, wang2025datawhisperer,xu2025unseen} that directly select data points from original datasets, DD methods are designed to generate novel data points through learning. The utility of DD methods has been witnessed in fields such as privacy protection \cite{chung2023rethinking, loo2023understanding, chen2022private, dong2022privacy}, continual learning \cite{rosasco2021distilled, gu2023summarizing, yang2023efficient, masarczyk2020reducing}, and neural architecture search \cite{bohdal2020flexible, medvedev2021learning, such2020generative}

Among the various DD techniques, matching-based methods, particularly gradient matching (GM) \cite{zhao2020dataset, zhao2021dataset, lee2022dataset, kim2022dataset} and trajectory matching (TM) \cite{cazenavette2022dataset, cui2023scaling, du2023minimizing, guo2023towards}, have demonstrated outstanding performance. However, a gap remains between their theoretical understanding and empirical success. To offer a unified explanation of these methods, we aim to explore the following question:

\textbf{Question 1:} \textit{Is there a unified theory to explain existing matching-based DD methods?}

\begin{figure*}[tb!]
    \centering
    \includegraphics[width=0.99\textwidth]{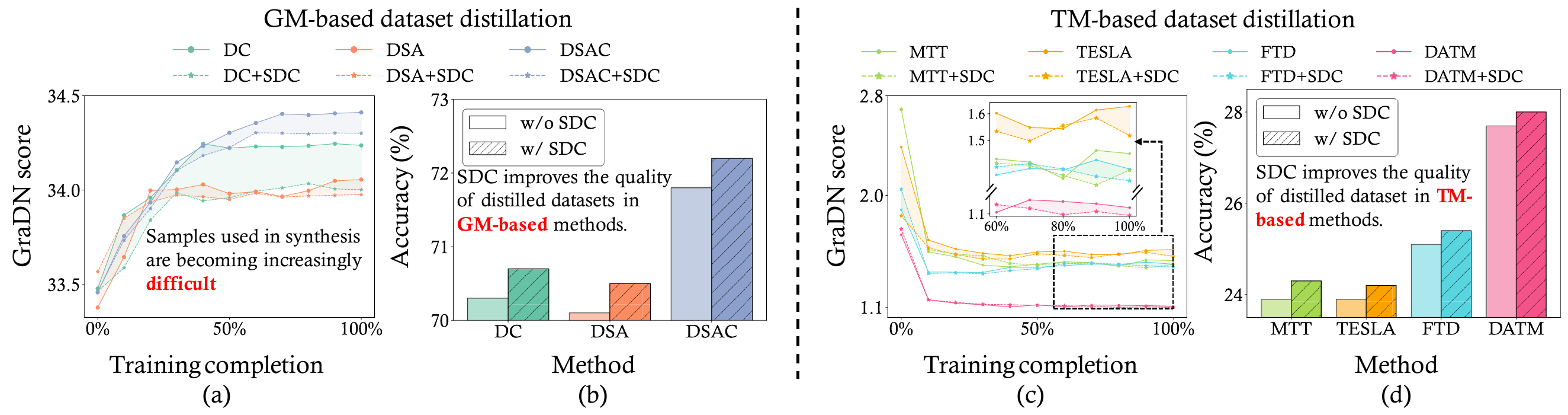}
    \vspace{-5pt}
    \caption{We conducted experiments of GM-based methods on the FashionMNIST dataset and TM-based methods on the CIFAR-100 dataset. (a) Average gradient norms of network parameters for different GM-based methods are enhanced during distillation. The shade represents the gap before and after adding our {\mymethod}. (b) Test performance w/ and w/o our proposed {\mymethod} on different GM-based methods. Our {\mymethod} are incorporated to improve all matching-based methods. (c) Average gradient norms of network parameters for different TM-based methods are alleviated during distillation. (d) Test performance w/ and w/o our proposed {\mymethod} on different TM-based methods. Note that the average gradient norms are smoothed using the exponential moving average. Best viewed in color. }
    \vspace{-10pt}
    \label{fig:Norm}
\end{figure*}

\begin{figure*}[tb!]
    \centering
    \includegraphics[width=0.99\textwidth]{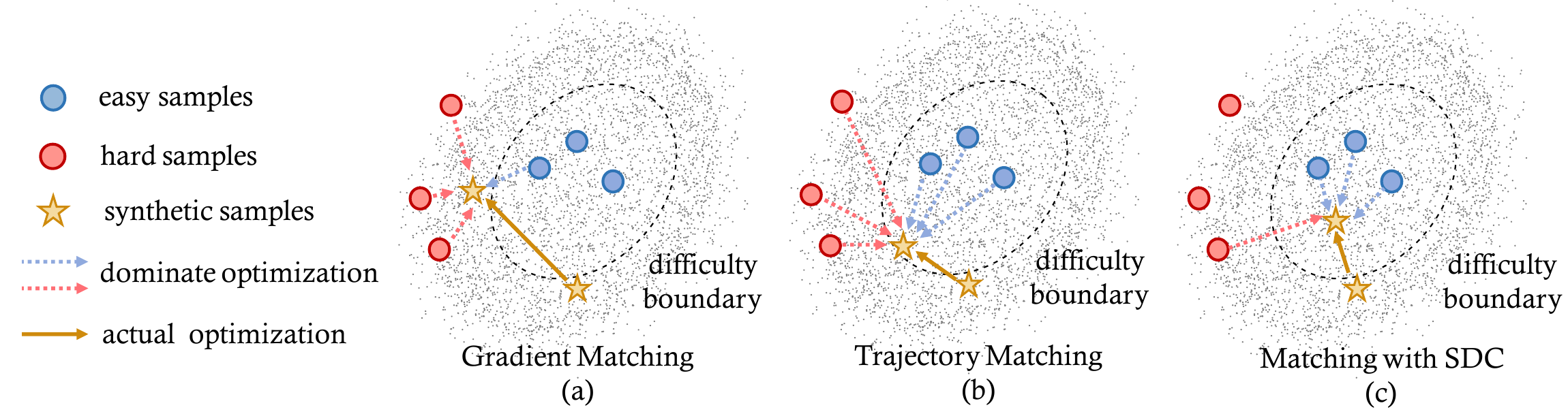}
    \caption{Comparison of matching-based dataset distillation methods from a sample difficulty perspective. (a) Gradient matching-based methods mainly utilize \hard{} samples during synthesizing. (b) Trajectory matching-based methods do not explicitly take the difficulty of samples into consideration. (c) Our {\mymethod} encourages matching-based methods to prioritize the synthesis of \easy{} samples.}
    \vspace{-10pt}
    \label{fig:overview}
\end{figure*}

To address Question 1, we first empirically examine the differences between matching-based distillation methods. It is widely acknowledged that sample difficulty (Definition~\ref{def:sample-difficulty}) is a crucial metric in data-centric AI that significantly affects model performance, as seen in dataset pruning \cite{tan2024data, maroto2024puma, maharana2023d2, sorscher2022beyond}, and large language model prediction \cite{cui2024learning, lin2024data, li2024selective}. To track the differences between current distillation methods, we follow \cite{paul2023deep} and analyze sample difficulty using the GraDN metric (Definition~\ref{def:gradn}). Surprisingly, we discover that the GraDN score is increased in GM-based methods (Figure~\ref{fig:Norm}(a)), while TM-based methods may reduce this metric (Figure~\ref{fig:Norm}(c)). These distinct trends indicate that the difficulty of samples utilized in GM-based methods is elevated (Figure~\ref{fig:overview}(a)), whereas in TM-based methods, it is reduced (Figure~\ref{fig:overview}(b)) during the distillation process.

Motivated by these observations, we develop a theoretical explanation for current DD methods from the perspective of sample difficulty. Specifically, we draw upon the \textit{neural scaling law} in the data pruning theory~\cite{sorscher2022beyond} to connect sample difficulty with performance. As shown in Figure~\ref{fig:theory}(c), our theory indicates that in matching-based DD methods, when the synthetic dataset is small—specifically, when the images-per-class (IPC) is low—the optimal strategy is to primarily focus on easier samples rather than harder ones to enhance performance. Based on our theory, we further explain why TM-based methods usually outperform GM-based methods in real scenarios.

Beyond developing a theoretical framework, we take steps to explore solutions for improving current approaches. This raises another key research question:

\textbf{Question 2:} \textit{Is it empirically feasible to identify a loss function that surpasses the performance of the matching loss by controlling the difficulty of learned patterns during distillation?}

To address Question 2, based on our empirical observations and theoretical analysis, we propose the novel \textit{\underline{\textbf{S}}ample \underline{\textbf{D}}ifficulty \underline{\textbf{C}}orrection} ({\mymethod}) method to improve the synthetic dataset quality in current matching-based distillation methods. We do this by guiding the distillation method to focus more on \easy{} samples than \hard{} samples, adding an implicit gradient norm regularizer to enhance quality.

Our contributions are listed as follows: 
\begin{itemize}[leftmargin=10pt, topsep=0pt, itemsep=1pt, partopsep=1pt, parsep=1pt] 
\item We empirically investigate \textit{sample difficulty} from the perspective of gradient norm in distillation methods, linking it to synthetic dataset quality. We propose that GM-based methods focus on difficult samples during optimization, while TM-based methods show no dominant preference for difficulty. This may explain the poorer performance of GM-based methods than TM-based methods.
\item We theoretically elucidate the mechanism of matching-based DD methods from the perspective of sample difficulty. Adapting the neural scaling law theory from data pruning \cite{sorscher2022beyond} to distillation settings, we provide insights into how matching strategies evolve with the size of the synthetic dataset. Consequently, we propose that focusing on matching \easy{} samples is a better strategy when the synthetic dataset is small. 
\item We introduce \textit{Sample Difficulty Correction} (\mymethod) to improve the quality of synthetic datasets in current matching-based DD methods. Our method demonstrates superior generalization performance across 7 distillation methods (DC \cite{zhao2020dataset}, DSA \cite{zhao2021dataset}, DSAC \cite{lee2022dataset}, MTT \cite{cazenavette2022dataset}, FTD \cite{du2023minimizing}, TESLA \cite{cui2023scaling}, DATM \cite{guo2023towards}) and 6 datasets (MNIST \cite{deng2012mnist}, FashionMNIST \cite{xiao2017fashionmnist}, SVHN \cite{netzer2011reading}, CIFAR-10/100 \cite{krizhevsky2009learning}, and Tiny-ImageNet \cite{le2015tiny}). \end{itemize}

\section{Preliminaries and Related Work}

Dataset distillation involves synthesizing a small, condensed dataset $\mathcal{D}_{\mathsf{syn}}$ that efficiently encapsulates the informational essence of a larger, authentic dataset $\mathcal{D}_{\mathsf{real}}$. %We add subscript to represent the network \textit{only} trained on $\mathcal{D}_{\mathsf{real}}$ as $\theta_t^{\mathcal{D}_{\mathsf{real}}}$, and $\mathcal{D}_{\mathsf{syn}}$ as $\theta_t^{\mathcal{D}_{\mathsf{syn}}}$.

\textbf{Gradient Matching} (GM) based methods are pivotal in achieving distillation by ensuring the alignment of training gradients between surrogate models trained on both the original dataset $\mathcal{D}_{\mathsf{real}}$ and the synthesized dataset $\mathcal{D}_{\mathsf{syn}}$. This method is first introduced by DC \cite{zhao2020dataset}. Let $\theta_t$ represent the network parameters sampled from distribution $P_\theta$ at step $t$, and $C$ symbolizes the categories within $\mathcal{D}_{\mathsf{real}}$. The cross-entropy loss $\mathcal{L}$, is employed to assess the matching loss by comparing the gradient alignment over a time horizon of $T$ steps. The formal optimization objective of DC is:
\begin{equation}
\label{eq:DC}
\underset{\mathcal{D}_{{\color{RoyalBlue} \mathsf{syn}}}}{\operatorname{\arg\min}} \underset{\theta_0 \sim P_\theta, c \sim C}{\mathbb{E}}
\left[ \sum_{t=0}^T \mathbf{D}\left(\nabla_\theta \mathcal{L}_{\mathcal{D}_{{\color{OrangeRed} \mathsf{real}}}^c}\left(\theta_t\right), \nabla_\theta \mathcal{L}_{\mathcal{D}_{{\color{RoyalBlue} \mathsf{syn}}}^c}\left(\theta_t\right)\right) \right],
\end{equation}

where $\mathbf{D}$ measures the cumulative distances (\emph{e.g.}, cosine/$L_2$ distance in DC) between the gradients of weights corresponding to each category output. The parameter updates for $\theta$ are executed in an inner loop via gradient descent, with a specified learning rate $\eta$:
\begin{equation}
    \theta_{t+1} \leftarrow \theta_t - \eta \cdot \nabla_\theta \mathcal{L}_{\mathcal{D}_{{\color{RoyalBlue} \mathsf{syn}}}}\left(\theta_t\right).
\end{equation}

Building upon this, DSA \cite{zhao2021dataset} enhances DC by implementing consistent image augmentations on both $\mathcal{D}_{\mathsf{real}}$ and $\mathcal{D}_{\mathsf{syn}}$ throughout the optimization process. Moreover, DCC \cite{lee2022dataset} refines the gradient matching objective by incorporating class contrastive signals at each gradient matching step, which results in enhanced stability and performance. Combining DSA and DCC, DSAC \cite{lee2022dataset} further introduces improvements by synergizing these techniques. The revised optimization objective for DCC and DSAC is formulated as:
% \begin{equation}
% \label{eq:DCC}
% \underset{\mathcal{D}_{{\color{RoyalBlue} \mathsf{syn}}}}{\operatorname{\arg\min}} \underset{\theta_0 \sim P_\theta}{\mathbb{E}}
% \left[ \sum_{t=0}^{T} \mathbf{D}\left( \mathbb{E}_{c \in C} \left[ \nabla_{\theta} \mathcal{L}_{\mathcal{D}_{{\color{OrangeRed} \mathsf{real}}}^{c}}(\theta_t) \right], 
% \mathbb{E}_{c \in C} \left[ \nabla_{\theta} \mathcal{L}_{\mathcal{D}^{c}_{{\color{RoyalBlue} \mathsf{syn}}}}(\theta_{t}) \right] \right) \right].
% \end{equation}
\begin{equation}
\label{eq:DCC}
\begin{split}
\underset{\mathcal{D}_{{\color{RoyalBlue} \mathsf{syn}}}}{\operatorname{\arg\min}} \underset{\theta_0 \sim P_\theta}{\mathbb{E}}
\left[ \sum_{t=0}^{T} \mathbf{D}\left( 
\mathbb{E}_{c \in C} \left[ \nabla_{\theta} \mathcal{L}_{\mathcal{D}_{{\color{OrangeRed} \mathsf{real}}}^{c}}(\theta_t) \right], \right. \right. \\
\left. \left. \mathbb{E}_{c \in C} \left[ \nabla_{\theta} \mathcal{L}_{\mathcal{D}^{c}_{{\color{RoyalBlue} \mathsf{syn}}}}(\theta_{t}) \right] \right) \right] .
\end{split}
\end{equation}

\textbf{Trajectory matching} (TM) based approaches aim to match the training trajectories of surrogate models by optimizing over both the real dataset $\mathcal{D}_{\mathsf{real}}$ and the synthesized dataset $\mathcal{D}_{\mathsf{syn}}$. TM-based methods were initially proposed in MTT \cite{cazenavette2022dataset}. Let term $\tau^{\mathcal{D}_{\mathsf{real}}}$ denote the expert training trajectories, represented as a sequential array of parameters $\{\theta_t^\mathcal{D_{\mathsf{real}}}\}_{t=0}^{T}$, obtained from training a network on the real dataset $\mathcal{D}_{\mathsf{real}}$. In parallel, ${\theta}_t^{\mathcal{D}_{\mathsf{syn}}}$ refers to the parameter set of the network trained on $\mathcal{D}_{\mathsf{syn}}$ at step $t$. In each iteration, parameters $\theta_t^{\mathcal{D}_{\mathsf{real}}}$ and $\theta_{t+M}^{\mathcal{D}_{\mathsf{real}}}$ are randomly selected from the expert trajectory pool $\{\tau^{\mathcal{D}_{\mathsf{real}}}\}$, serving as the initial and target parameters for trajectory alignment, where $M$ is a predetermined hyperparameter. TM-based methods enhance the synthetic dataset $\mathcal{D}_{\mathsf{syn}}$ by minimizing the loss defined as:
\vspace{-5pt}
\begin{equation}
\label{eq:TM}
\underset{\mathcal{D}_{{\color{RoyalBlue} \mathsf{syn}}}}{\arg\min} \underset{\theta_0 \sim P_\theta}{\mathbb{E}}\left[\sum_{t=0}^{T-M} \frac{\mathbf{D}\left(\theta_{t+M}^{\mathcal{D}_{{\color{OrangeRed} \mathsf{real}}}}, \theta_{t+N}^{\mathcal{D}_{{\color{RoyalBlue} \mathsf{syn}}}}\right)}{\mathbf{D}\left(\theta_{t+M}^{\mathcal{D}_{{\color{OrangeRed} \mathsf{real}}}}, \theta_t^{\mathcal{D}_{{\color{OrangeRed} \mathsf{real}}}}\right)}\right],
\end{equation}

where $\mathbf{D}$ is a distance metric (\emph{e.g.}, $L_2$ distance in MTT) and $N<<M$ is a predefined hyperparameter. ${\theta}_{t+N}^{\mathcal{D}_{\mathsf{syn}}}$ is derived through an inner optimization using the cross-entropy loss $\mathcal{L}$ with the learning rate $\eta$:
\vspace{-5pt}
\begin{equation}
{\theta}_{t+i+1}^{\mathcal{D}_{{\color{RoyalBlue} \mathsf{syn}}}} \leftarrow {\theta}_{t+i}^{\mathcal{D}_{{\color{RoyalBlue} \mathsf{syn}}}} - \eta \nabla_{\theta} \mathcal{L}_{\mathcal{D}_{{\color{RoyalBlue} \mathsf{syn}}}}({\theta}_{t+i}^{\mathcal{D}_{{\color{RoyalBlue} \mathsf{syn}}}}), \text{ where } {\theta}_t^{\mathcal{D}_{{\color{RoyalBlue} \mathsf{syn}}}} := \theta_t^{\mathcal{D}_{{\color{OrangeRed} \mathsf{real}}}}.
\end{equation}

Similarly, TESLA \cite{cui2023scaling} utilizes linear algebraic manipulations and soft labels to increase compression efficiency, FTD \cite{du2023minimizing} aims to seek a flat trajectory to avoid accumulated trajectory error, and DATM \cite{guo2023towards} considers matching only necessary parts of trajectory with difficulty alignment.

% \textbf{Connection between GM-based and TM-based methods.} In fact, GM-based methods can be approximately regarded as a special case of TM-based methods when $M=N=1$ under discrete optimization scenario. Empirically, TM-based methods smooth the long trajectory generated by multiple network checkpoints, while GM-based methods take a single step in such long trajectory.

\section{Method}

\begin{figure*}[tb!]
    \centering
    \includegraphics[width=0.99\textwidth]{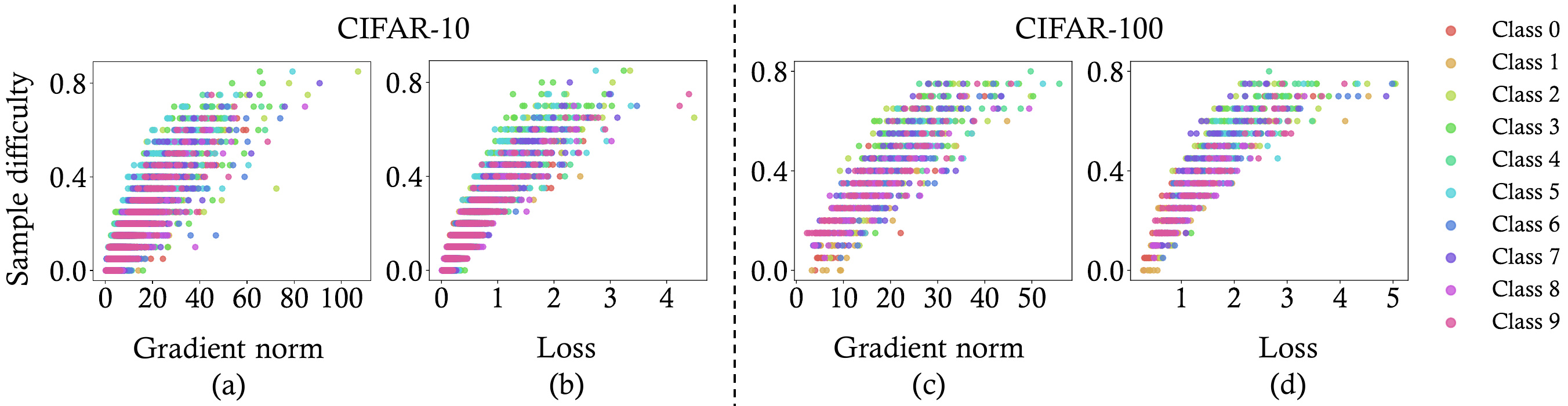}
    \vspace{10pt}
    \caption{The statistical relationship between sample difficulty $\chi(x,y;\Theta_t)$, gradient norm $\text{GraDN}(x,y;\Theta_t)$, and average validation loss for each sample $(x,y)$ on a series of models with $\theta_t\in \Theta_t$. We observe a significant positive correlation between sample difficulty and both the gradient norm and the loss. Experiments were conducted using ResNet-18 on CIFAR-10 and ResNet-34 on CIFAR-100. Each score was evaluated across 20 pretrained models. For CIFAR-100, 10 categories were randomly selected for visualization. The relationships depicted are: (a) sample difficulty vs. gradient norm on CIFAR-10, (b) sample difficulty vs. loss on CIFAR-10, (c) sample difficulty vs. gradient norm on CIFAR-100, and (d) sample difficulty vs. loss on CIFAR-100.}
    \label{fig:difficulty}
    % \vspace{-5pt}
\end{figure*}

\subsection{A Closer Look at Sample Difficulty}
\label{sec:empirical}
In this subsection, we aim to intuitively understand dataset distillation through the concept of sample difficulty (Definition~\ref{def:sample-difficulty}), which is pivotal in data-centric AI \cite{tan2024data, maroto2024puma, maharana2023d2, choi2021vab, xie2020deal,cui2024learning,li2024selective}. We begin by empirically observing the evolution of sample difficulty during the distillation process. Firstly, we introduce the commonly used definition of sample difficulty, namely the GraDN score (Definition~\ref{def:gradn}), and validate the reliability of this metric. Furthermore, we track the GraDN score across current dataset distillation methods to delve deeper into their underlying mechanisms.

\begin{definition}[Sample Difficulty \cite{meding2022trivial}]
\label{def:sample-difficulty}
    Given a training pair $(x, y)$ and a series of pretrained models at training time $t$, the sample difficulty, denoted $\chi(x, y;\Theta_t)$, is defined as the expected probability of $(x, y)$ being misclassified by an ensemble of models $\theta_t \in \Theta_t$. Formally, it is presented as:
    \begin{equation}
    \chi(x, y;\Theta_t) = \mathbb{E}_{\theta_t \in \Theta_t}\left[\mathbbm{1}\left(y\neq \theta_t(x)\right)\right],
    \end{equation}
    where $\mathbbm{1}\left(z\right)$ is an indicator function that equals 1 if the boolean input $z$ is true, and 0 otherwise. In this case, the indicator function equals to 1 if the sample $(x,y)$ is misclassified by the model with parameters $\theta_t$, and 0 otherwise.
\end{definition}

\begin{definition}[GraDN Score \cite{paul2023deep}]
\label{def:gradn}
Consider a training pair $(x, y)$, with $\mathcal{L}$ representing the loss function. At time $t$, the GraDN score for $(x, y)$ is calculated as the average gradient norm of the loss $\mathcal{L}$ across a diverse ensemble of models with parameters $\theta_t\in \Theta_t$:
\begin{equation}
\operatorname{GraDN}(x, y;\Theta_t) = \mathbb{E}_{\theta_t\in \Theta_t}\left[\|\nabla_{\theta} \mathcal{L}(x, y;\theta_t)\|_2\right],
\end{equation}
where $\nabla_{\theta} \mathcal{L}(x, y;\theta_t)$ denotes the gradient of loss $\mathcal{L}$ on sample $(x,y)$ \emph{w.r.t.} the model parameters $\theta_t$, and $\|\cdot\|_2$ denotes $L_2$ norm.
\end{definition}

According to \cite{meding2022trivial}, the difficulty of each sample can be assessed by the misclassification ratio across a series of pretrained models (Definition~\ref{def:sample-difficulty}). Additionally, from an optimization perspective, it can be represented by the gradient norm of the loss on a series of pretrained models for this sample (Definition~\ref{def:gradn}). In our study, we adopt Definition~\ref{def:gradn} to evaluate sample difficulty as interpreted by various matching methods.

\textbf{Empirical verification of the relationship between sample difficulty and gradient norm.} We conducted experiments to verify the reliability of the GraDN score in classifying the CIFAR-10 and CIFAR-100 datasets by training a set of models. As depicted in Figure~\ref{fig:difficulty}, the GraDN score shows a clear positive correlation with the sample difficulty. For easier samples, GraDN scores are generally lower, exerting minimal impact on the network's gradient flow. Conversely, for harder samples, higher GraDN scores indicate a significant impact on the optimization directions of the models. We show detailed results of the relationships between these metrics in Appendix~\ref{supp:difficulty}. 

\textbf{Exploring sample difficulty across different distillation methods.} Beyond sample difficulty under classification scenarios, we now extend our observations to matching-based distillation methods. Specifically, we examined the average gradient norm of the training cross-entropy loss across network parameters during the distillation process. As shown in Figure~\ref{fig:Norm}(a)(c), we found that the average gradient norm (corresponding to the GraDN score) tends to increase in GM-based methods (signifying harder samples), whereas it decreases in TM-based methods (indicating easier samples). This unexpected phenomenon motivates us to further theoretically explore matching-based distillation methods from the perspective of sample difficulty.

\subsection{An Analytical Theory for Explaining Matching-based Dataset Distillation}
\label{sec:theory}

In Section \ref{sec:empirical}, we empirically observed distinct trends in \textit{sample difficulty} across various dataset distillation methods. Here, we propose an analytical theory based on the \textit{neural scaling law} to formally analyze sample difficulty in matching-based methods. We extend the theory of data pruning presented by \cite{sorscher2022beyond} and validate its applicability within the context of DD using an expert-student perceptron model. Unlike data pruning, where the pruned dataset is directly selected from the original dataset, DD involves synthesizing a small, new, unseen dataset.

\begin{figure*}[tb!]
    \centering
    {\includegraphics[width=0.99\textwidth]{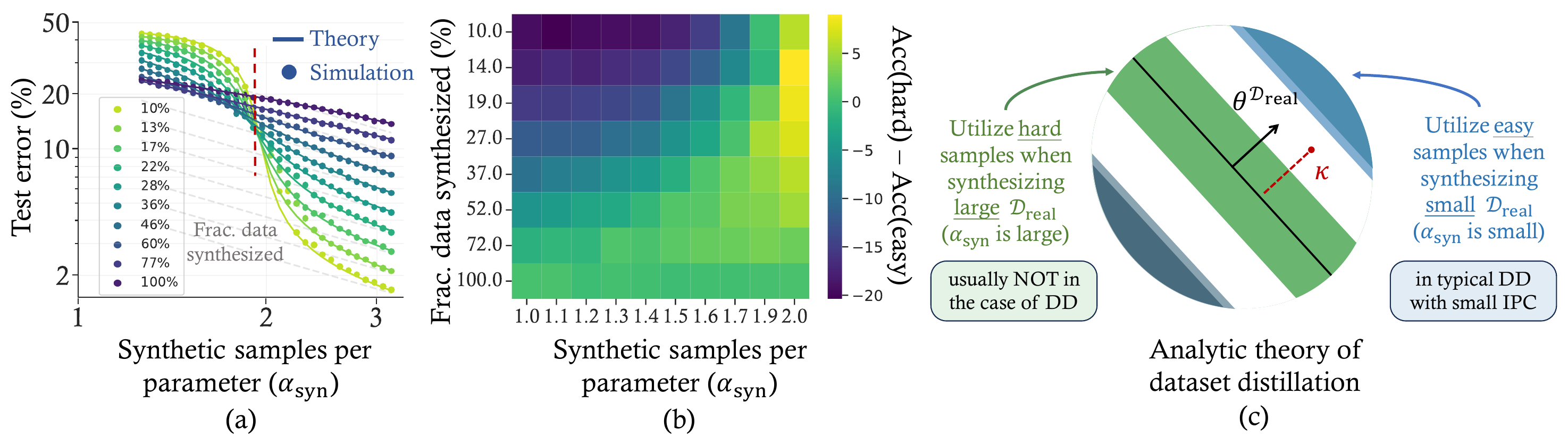}}
    \vspace{-5pt}
    \caption{An analytic theory of dataset distillation. (a) Test error $\varepsilon$ as a function of the synthetic samples per parameter $\alpha_{\mathsf{syn}}$ and fraction of data synthesized $f$  in a perfect expert setting ($\gamma=0$). (b) We show the difference in test accuracy in learning synthetic dataset by learning from \hard{} vs \easy{} samples, revealing the change in distillation strategy. (c) Our theory suggests that when we synthesize small dataset (small $\alpha_{\mathsf{syn}}$), the better distillation strategy is to utilize the \easy{} samples.}
    \label{fig:theory}
\vspace{-15pt}
\end{figure*}

We start our analysis with tools from statistical mechanics \cite{mezard1987spin}. Let us consider a classification problem in dataset $\mathcal{D}^{\mathsf{real}}$ containing $d_{\mathsf{real}}$ samples $\{x_i,y_i\}_{i=1,\ldots,d_{\mathsf{real}}}$, where $x_i \in \mathbb{R}^d \sim \mathcal{N}(0,I_{d})$ are \emph{i.i.d.} zero-mean, unit variance Gaussian inputs, and $y_i = \text{sign}(\theta^{\mathcal{D}_{\mathsf{real}}^\top} x_i) \in \{-1,+1\}$ are labels generated by an expert perceptron $\theta^{\mathcal{D}_{\mathsf{real}}} \in \mathbb{R}^{d}$. Our analysis is within the high-dimensional statistics limit, where $d, d_{\mathsf{real}} \to \infty$ while maintaining the ratio of total training samples to parameters $\alpha_{\mathsf{tot}} = d_{\mathsf{real}}/d$ at $O(1)$. The general distillation algorithm proceeds as follows:
\begin{enumerate}[leftmargin=15pt, topsep=0pt, itemsep=1pt, partopsep=1pt, parsep=1pt]
    \item Train a student perceptron on $\mathcal{D}_{\mathsf{real}}$ for a few epochs to obtain weights $\theta^{\mathsf{probe}}$. The gap between can be measured by the angle $\gamma$ between the probe student $\theta^{\mathsf{probe}}$ and the expert $\theta^{\mathsf{real}}$. If $\theta^{\mathsf{probe}}\approx \theta^{\mathsf{real}}$, we denote the $\theta^{\mathsf{probe}}$ as a \textit{perfect} probe ($\gamma=0$). Otherwise, in \textit{imperfect} probe cases, $\gamma\neq 0$.
    \item Compute the margin $m_i = \theta^{\mathsf{probe}^\top} (y_i x_i)$ for each training example, categorizing large (small) margins as \easy{} (\hard{}) samples.
    \item Generate a synthetic dataset $\mathcal{D}_{\mathsf{syn}}$ of size $d_{\mathsf{syn}} = f d_{\mathsf{real}}$, by learning from the hardest samples from $\mathcal{D}_{\mathsf{real}}$ for few steps. The ratio of total samples of the synthetic dataset to parameters is $\alpha_{\mathsf{syn}} = d_{\mathsf{syn}}/d$.
    \item Train a new perceptron $\theta^{\mathcal{D}_{\mathsf{syn}}}$ on $\mathcal{D}_{\mathsf{syn}}$, aiming to classify training data with the maximal margin $\kappa = \min_i \theta^{\mathcal{D}_{\mathsf{syn}}^\top} y_ix_i$.
\end{enumerate}

We analyze the test error $\varepsilon$ of this final perceptron $\theta^{\mathcal{D}_{\mathsf{syn}}}$ as a function of $\alpha_\mathsf{syn}$, $f$, and the angle $\gamma$ between the probe student $\theta^{\mathsf{probe}}$ and the expert $\theta^{\mathcal{D}_{\mathsf{real}}}$.  Similar to \cite{sorscher2022beyond}, we define $\theta^{\mathsf{probe}}$ as a random Gaussian vector conditioned to have angle $\gamma$ with the expert. Under this scenario, we can derive an asymptotically exact analytic expression for $\varepsilon(\alpha_\mathsf{syn},f,\gamma)$ (see Appendix \ref{supp:theory} for details):
\vspace{-5pt}
\begin{equation}
    \label{eq:generalization_error}
    \varepsilon = \frac{\cos^{-1}(R)}{\pi}, \quad \text{where } R=\frac{\theta^{\mathcal{D}_{{\color{OrangeRed} \mathsf{real}}}^\top} \theta^{\mathcal{D}_{{\color{RoyalBlue} \mathsf{syn}}}}}{\|\theta^{\mathcal{D}_{{\color{OrangeRed} \mathsf{real}}}}\|_2 \|\theta^{\mathcal{D}_{{\color{RoyalBlue} \mathsf{syn}}}}\|_2}
\end{equation}

Likewise in \cite{sorscher2022beyond}, we can solve $R$ with saddle point equations in Appendix~\ref{supp:perfect}, enabling direct predictions of the test error $\varepsilon$ according to Eq~(\ref{eq:generalization_error}).

\textbf{Verification of the neural scaling law in dataset distillation}.  
We first evaluated the correctness of our theory in the perfect expert-student setting ($\gamma=0$). As shown in Figure \ref{fig:theory}(a), we observed an excellent match between our analytic theory (solid curves) and numerical simulations (dots) of perceptron learning at parameters $d=200$ in dataset distillation. We also verify our theory in imperfect probe settings when $\gamma\neq 0$, as shown in  Appendix~\ref{supp:imperfect}.

% \wsb{
% \begin{equation}
% R=\frac{2\alpha}{f\sqrt{2\pi}\sqrt{1-R^{2}}}\int_{-\infty}^{\kappa}Dt\ \exp\bigg(-\frac{R^{2}t^{2}}{2(1-R^{2})}\bigg)\bigg[1-\exp\bigg(-\frac{\gamma(\gamma-2Rt)}{2(1-R^{2})}\bigg)\bigg](\kappa-t) 
% \end{equation}

% \begin{equation}
% 1-R^{2}=\frac{2\alpha}{f}\int_{-\infty}^{\kappa}Dt\ \bigg[H\bigg(-\frac{Rt}{\sqrt{1-R^{2}}}\bigg)-H\bigg(-\frac{Rt-\gamma}{\sqrt{1-R^{2}}}\bigg)\bigg](\kappa-t)^{2}
% \end{equation}

% where
% \begin{equation}
% H(x) = \frac{1}{2} \left(1 - \frac{2}{\sqrt{\pi}} \int_0^{\frac{x}{\sqrt{2}}} e^{-t^2} \, dt\right)
% \end{equation}
% }

\textbf{The neural scaling law for dataset distillation}.  We further investigate the relationship between the distillation ratio ($\alpha_{\mathsf{syn}}$), the fraction of data synthesized ($f$), and the final accuracy of $\theta^{\mathcal{D}_\mathsf{syn}}$ under various distillation strategies, such as synthesizing $\mathcal{D}_{\mathsf{syn}}$ from \hard{} or \easy{} samples. Similar to data pruning, when $f = 1$ (no distillation), the test error follows the classical perceptron learning power-law scaling, $\varepsilon \propto \alpha_\mathsf{syn}^{-1}$. In other cases, our findings reveal that for smaller $\alpha_\mathsf{syn}$ (smaller synthetic datasets), learning from the \hard{} samples results in poorer performance than no distillation. Conversely, for larger $\alpha_\mathsf{syn}$, focusing on the \hard{} samples yields substantially better outcomes than no distillation. We find that in limited data regimes, matching the \easy{} samples, which have the largest margins, offers a more effective distillation strategy. This finding highlights that in most cases of DD (where $d_{\mathsf{syn}} << d_{\mathsf{real}}$), it is crucial for the model to first learn from the basic characteristics in $\mathcal{D}_{\mathsf{real}}$; hence, prioritizing \easy{} samples facilitates reaching a moderate error level more swiftly.

\textbf{Understanding GM-based and TM-based methods with the neural scaling law.} As depicted in Figure~\ref{fig:Norm}(a), we observe that GM-based methods typically incorporate the \hard{} samples within the synthetic dataset. This trend is due to the GM-based matching loss in Eq.(~\ref{eq:DC}), which predominantly penalizes samples with large gradients (\hard{} samples) (shown in Figure~\ref{fig:overview}(b)). However, in most DD settings, the size of synthetic dataset $\mathcal{D}_{\mathsf{syn}}$ is usually small. Therefore, according to our theory, we should mainly focus on synthesizing the dataset by matching  \easy{} samples to achieve higher dataset quality. In contrast, the simplicity of the synthetic samples generated by TM-based methods, as shown in Figure~\ref{fig:Norm}(a), is not directly concerned through distillation. From Eq.(~\ref{eq:TM}), it is evident that TM-based methods prioritize parameter alignment, thus penalizing the matching term without explicitly targeting sample difficulty (shown in Figure~\ref{fig:overview}(c)). This approach results in a synthetic dataset that may be generated by learning samples of randomly vary in difficulty. We can provide a explanation that TM-based methods generalize well in real scenarios than GM-based methods because of they do not explicitly focus on synthesizing by matching \hard{} samples. 

% We propose our Proposition~\ref{conclusion} as follows:
% \begin{proposition}
% \label{conclusion}
% In general, in order to learn a synthetic dataset $\mathcal{D}_{\mathsf{syn}}$ with size $d_{\mathsf{syn}}$ from $\mathcal{D}_{\mathsf{real}}$ with size $d_{\mathsf{real}}$ (where $d_{\mathsf{syn}} \ll d_{\mathsf{real}}$), \easy{} samples from $\mathcal{D}_{\mathsf{real}}$ should be primarily utilized to enhance the quality of $\mathcal{D}_{\mathsf{syn}}$.
% \end{proposition}

\subsection{Matching with Sample Difficulty Correction}

Based on our theoretical analysis of matching-based dataset distillation, we propose a novel method to enhance existing techniques for synthesizing higher-quality distilled datasets. Although TM-based methods have achieved relative success on current benchmark datasets, they do not explicitly consider sample difficulty, which could ensure higher synthetic dataset quality.

A direct approach to impose constraints on sample difficulty is to calculate the gradient norm for each sample as a metric to determine its utility. Let us consider the case of GM-based methods. At step $t$, a batch of real samples $\mathcal{B}_{\mathsf{real}}^c \sim \mathcal{D}_{\mathsf{real}}^c$ of class $c \in C$ is to be matched with the gradients of a synthetic batch $\mathcal{B}_{\mathsf{syn}}^c \sim \mathcal{D}_{\mathsf{syn}}^c$. To decide whether to utilize each sample in $\mathcal{B}_{\mathsf{real}}^c$, it is natural to compute the gradient norm of each sample and utilize those with a score smaller than a predefined threshold $\tau$. Specifically, a sample $(x, y)$ is utilized if $\|\nabla_{\theta}\mathcal{L}(x, y; \theta_t)\|_2 \leq \tau$. Consequently, the modified loss for matching only \easy{} samples is:
\begin{equation}
    \label{eq:naive-SDC}
    \begin{aligned}
    & \mathcal{L}_{\tilde{\mathcal{B}}_{{\color{OrangeRed}\mathsf{real}}}^c}= \mathbb{E}_{(x,y)\in \tilde{\mathcal{B}}_{{\color{OrangeRed}\mathsf{real}}}^c} \left[\mathcal{L}(x,y;\theta_t)\right], \\ & \mathcal{L}_{\tilde{\mathcal{B}}_{{\color{RoyalBlue}\mathsf{syn}}}^c}= \mathbb{E}_{(x,y)\in \tilde{\mathcal{B}}_{{\color{RoyalBlue}\mathsf{syn}}}^c} \left[\mathcal{L}(x,y;\theta_t)\right],
    \end{aligned}
\end{equation}
where $\tilde{\mathcal{B}}_{\mathsf{real}}^c = \{(x,y)| (x,y)\in \mathcal{B}_{\mathsf{real}}^c, \|\nabla_{\theta} \mathcal{L}(x, y;\theta_t)\|_2 \leq \tau \}$ denotes the modified batch with only \easy{} samples, and $\tilde{\mathcal{B}}_{\mathsf{syn}}^c$ denote a sampled batch from $\mathcal{D}_{\mathsf{syn}}^c$ with the same size as $\tilde{\mathcal{B}}_{\mathsf{real}}^c$. The corresponding matching loss should be:
\begin{equation}
    \tilde{L}(\theta_t)=\mathbf{D}\left(\nabla_\theta \mathcal{L}_{\tilde{\mathcal{B}}_{{\color{OrangeRed}\mathsf{real}}}^c}\left(\theta_t\right), \nabla_\theta \mathcal{L}_{\tilde{\mathcal{B}}_{{\color{RoyalBlue}\mathsf{syn}}}^c}\left(\theta_t\right)\right),
\end{equation} 
However, the computational cost of constructing reduced \easy{} sample batch $\tilde{\mathcal{B}}_{\mathsf{real}}^c$ from $\mathcal{B}_{\mathsf{real}}^c$ is unrealistic in real-world scenarios because it requires calculating the gradient norm for each sample independently, resulting in a tenfold or greater increase in time. Besides, determining the difficulty threshold $\tau$ is also ad-hoc and challenging for each sample. Therefore, we take an alternative approach, \emph{i.e.}, we consider adding the overall sample difficulty of the whole batch $\mathcal{B}_{\mathsf{syn}}^c$ as an implicit regularization term in the matching loss function. Our proposed methods, named \textit{Sample Difficulty Correction} ({\mymethod}), can be incorporated into current matching methods with minimal adjustment of code implementation. Specifically, for a single-step GM, we have the following modified loss:
\begin{equation}
\begin{aligned}
    L_{\lambda}(\theta_t) & =\underbrace{\mathbf{D}\left(\nabla_\theta \mathcal{L}_{\mathcal{B}_{{\color{OrangeRed}\mathsf{real}}}^c}\left(\theta_t\right), \nabla_\theta \mathcal{L}_{\mathcal{B}_{{\color{RoyalBlue}\mathsf{syn}}}^c}\left(\theta_t\right)\right)}_{\text{Gradient Matching Loss}} \\
    &+ \underbrace{\lambda \left\|\nabla_\theta \mathcal{L}_{\mathcal{B}_{{\color{RoyalBlue}\mathsf{syn}}}^c}\right\|_2}_{\text{Gradient Norm Regularization}} 
    \end{aligned}
\end{equation}
For TM-based methods that do not explicitly focus on sample difficulty during distillation, we compute the average gradient norm of the whole dataset $\mathcal{D}_{\mathsf{syn}}$ during the optimization of the student network $\theta_{t}^{\mathcal{D}_{\mathsf{syn}}}$ \emph{w.r.t.} the training loss as the regularization term. Specifically, we have:
\begin{equation}
\begin{aligned}
     L_{\lambda}(\theta^{\mathcal{D}_{{\color{RoyalBlue}\mathsf{syn}}}}_t) &=
    \underbrace{\mathbf{D}\left(\theta_{t+M}^{\mathcal{D}_{{\color{OrangeRed}\mathsf{real}}}}, \theta_{t+N}^{\mathcal{D}_{{\color{RoyalBlue}\mathsf{syn}}}}\right) / {\mathbf{D}\left(\theta_{t+M}^{\mathcal{D}_{{\color{OrangeRed}\mathsf{real}}}}, \theta_t^{\mathcal{D}_{{\color{OrangeRed}\mathsf{real}}}}\right)}}_{\text{Trajectory Matching Loss}} \\
    &+ \underbrace{\lambda \left\|\nabla_\theta \mathcal{L}_{\mathcal{D}_{{\color{RoyalBlue}\mathsf{syn}}}}\right\|_2}_{\text{Gradient Norm Regularization}}
\end{aligned}
\end{equation}

By adding the gradient norm regularization, we can implicitly enforce current matching-based methods to mainly concentrate on synthesizing \easy{} samples to achieve better synthetic data quality. We provide the algorithm pseudocodes for GM- and TM-based methods in Appendix \ref{supp:algs}.

\begin{table*}[tb!]
  \caption{Comparison of test accuracy (\%) results of GM-based dataset distillation methods w/ and w/o SDC on MNIST, FashionMNIST, and SVHN datasets.}
  \label{table:GM}
  \centering
  \resizebox{0.9\textwidth}{!}{
  \begin{tabular}{c|ccc|ccc|ccc}
    \toprule
      Dataset & \multicolumn{3}{c|}{MNIST} & \multicolumn{3}{c|}{FashionMNIST} & \multicolumn{3}{c}{SVHN} \\
     IPC & 1 & 10 & 50 & 1 & 10 & 50 & 1 & 10 & 50 \\
     Ratio (\%)  & 0.02 & 0.2 & 1 & 0.2 & 2 & 10 & 0.2 & 2 & 10 \\
    \midrule
      Random & 64.9±3.5 & 95.1±0.9 & 97.9±0.2 & 51.4±3.8 & 73.8±0.7 & 82.5±0.7 & 14.6±1.6 & 35.1±4.1 & 70.9±0.9 \\
     Herding & 89.2±1.6 & 93.7±0.3 & 94.8±0.2 & 67.0±1.9 & 71.1±0.7 & 71.9±0.8 & 20.9±1.3 & 50.5±3.3 & 72.6±0.8 \\
     Forgetting & 35.5±5.6 & 68.1±3.3 & 88.2±1.2 & 42.0±5.5 & 53.9±2.0 & 55.0±1.1 & 12.1±1.7 & 16.8±1.2 & 27.2±1.5\\
    \midrule
    % DC  & 91.7±0.5 & 97.4±0.2 & 98.8±0.2 & 70.5±0.6 & 82.3±0.4 & 83.6±0.4 & 31.2±1.4 & 76.1±0.6 & 82.3±0.3\\
    DC  & 91.8±0.4 & 97.4±0.2 & 98.5±0.1 & 70.3±0.7 & 82.1±0.3 & 83.6±0.2 & 31.1±1.3 & 75.3±0.6 & 82.1±0.2\\
    % 76.1±0.5
    \textbf{{+\mymethod}}   & 92.0±0.4& 97.5±0.1 & 98.9±0.1  & 70.7±0.5 & 82.4±0.3 & 84.7±0.2& 31.4±1.2 & 76.0±0.5 & 82.3±0.3\\
    % \underline{76.1±0.5}
    \midrule
    % DSA  & 88.7±0.6 & 97.8±0.1 & 99.2±0.1 & 70.6±0.6 & 84.6±0.3 & 88.7±0.2 & 27.5±1.4 & 79.2±0.5 & 84.4±0.4\\
    DSA  & 88.9±0.8 & 97.2±0.1 & 99.1±0.1 & 70.1±0.4 & 84.7±0.2 & 88.7±0.2 & 29.4±1.0 & 79.2±0.4 & 84.3±0.4\\
    % 99.2±0.1
    %DCC  & 32.9±0.8 & 49.4±0.5 & 61.6±0.4 & 13.3±0.3 & 30.6±0.4 & 40.0±0.3 & - & - & -\\
    % \textbf{{+\mymethod}}  & 89.2±0.4& 97.9±0.1 & 99.2±0.4 & 70.9±0.5 & 84.8±0.2 &88.9±0.1 &28.9±1.0 & 79.4±0.4& 84.5±0.3\\
    \textbf{{+\mymethod}}  & 89.2±0.4& 97.3±0.1 & 99.2±0.4 & 70.5±0.5 & 84.8±0.2 &88.9±0.1 &30.6±1.0 & 79.4±0.4& 85.3±0.4\\
    % 99.2±0.4
    \midrule
    % DSAC  & 89.2±0.7 & 97.7±0.1 & 98.8±0.5 & 71.7±0.7 & 84.9±0.2 &87.8±0.2 & 47.5±2.6 & 80.5±0.6 & 87.2±0.3\\
    DSAC  & 89.2±0.7 & 97.7±0.1 & 98.8±0.1 & 71.8±0.7 & 84.9±0.2 &88.5±0.2 & 47.5±1.8 & 80.1±0.5 & 87.3±0.2\\
    % 98.9±0.1
    %DCC\textbf{{+\mymethod}}  & xxx & xxx & xxx & xxx & xxx & xxx & xxx & xxx & xxx\\
     % \textbf{{+\mymethod}}  &89.6±0.8  &97.7±0.2 &98.9±0.1 &72.2±0.6 &85.1±0.1 &88.8±0.2 &48.1±1.6&80.5±0.3 &87.3±0.2\\
     \textbf{{+\mymethod}}  &89.7±0.7  &97.8±0.1 &98.9±0.1 &72.2±0.6 &85.1±0.1 &88.7±0.1 &48.1±1.6&80.4±0.3 &87.4±0.2\\
     % \underline{98.9±0.1}
     \midrule
      Whole Dataset & \multicolumn{3}{c|}{99.6±0.0} & \multicolumn{3}{c|}{93.5±0.1} & \multicolumn{3}{c}{95.4±0.1} \\
    \bottomrule
  \end{tabular}
  }
  % \vspace{-10pt}
\end{table*}

\section{Experiments}\label{sec:exp}
\subsection{Basic Settings}\label{sec:setting}
\textbf{Datasets and baselines}. For GM-based methods, we followed previous works to conduct experiments on MNIST \cite{deng2012mnist}, FashionMNIST \cite{xiao2017fashionmnist}, SVHN \cite{netzer2011reading} datasets. We utilized current GM-based methods, including DC \cite{zhao2020dataset}, DSA \cite{zhao2021dataset}, and DSAC \cite{lee2022dataset} as baselines. For TM-based methods, we followed the recent papers to use CIFAR-10, CIFAR-100 \cite{krizhevsky2009learning}, and Tiny ImageNet \cite{le2015tiny} datasets. We performed experiments on current baselines including MTT \cite{cazenavette2022dataset}, FTD \cite{du2023minimizing}, TESLA \cite{cui2023scaling}, and DATM \cite{guo2023towards}. We added our \textit{Sample Difficulty Correction} (SDC) for all these baseline methods. To ensure a fair comparison, we employed identical hyperparameters for GM-based and TM-based methods with and without SDC while keeping all other variables constant, such as model architecture and augmentations. As per convention, for TM-based methods, we used max test accuracy, while for GM-based methods, we utilized the test accuracy from the last iteration.  We also compared our methods with classical data pruning algorithms including Random, Herding \cite{welling2009herding}, and Forgetting \cite{toneva2018empirical}. All hyperparameters are detailed in Appendix~\ref{supp:params}.

\textbf{Neural networks for distillation.}We used ConvNet as default to conduct experiments. Consistent with other previous methods, we used 3-layer ConvNet for CIFAR-10, CIFAR-100, MNIST, SVHN, and FashionMNIST, and 4-layer ConvNet for Tiny ImageNet. 

\subsection{Main Results}

\textbf{GM-based methods on MNIST, FashionMNIST, and SVHN.} As presented in Table~\ref{table:GM}, we report the results of three GM-based methods applied to MNIST, FashionMNIST, and SVHN datasets. Each method was evaluated with IPC (images-per-class) values of 1, 10, and 50. Notably, adding SDC improves the test accuracy of baseline methods across all datasets and IPC values, demonstrating the effectiveness of our approach. Notably, adding SDC to the original method improved the test accuracy of DSA by \textbf{1.2\%} on the SVHN dataset with IPC = 1, and by \textbf{1\%} with IPC = 50. For DC on the FashionMNIST dataset with IPC = 50, the test accuracy was increased by \textbf{1.1\%} with SDC. All hyperparameters are detailed in Table~\ref{tab:hyperparametersGM}. 

\textbf{TM-based methods on CIFAR-10/100 and Tiny ImageNet.} As shown in Table~\ref{table:TM}, we present the results of four TM-based methods trained on CIFAR-10, CIFAR-100 and Tiny ImageNet.  By incorporating the average gradient norm as a regularization term during matching with SDC, the resulting test accuracy was generally improved. Notably, employing SDC improved the test accuracy of FTD on CIFAR-10 by \textbf{1.2\%} with IPC = 10 and \textbf{1.1\%} with IPC = 50, and enhanced the test accuracy of DATM on Tiny ImageNet by \textbf{0.6\%}. For FTD, we used EMA (exponential moving average) just as in the original method\cite{du2023minimizing}. All hyperparameters are detailed in  Tables~\ref{tab:hyperparametersMTT},~\ref{tab:hyperparametersTESLA},~\ref{tab:hyperparametersFTD}, and~\ref{tab:hyperparametersDATM}.

\begin{table*}[tb!]
  \caption{Comparison of test accuracy (\%) results of TM-based dataset distillation methods w/ and w/o SDC on CIFAR-10, CIFAR-100, and Tiny ImageNet datasets.}
  \vspace{-5pt}
  \label{table:TM}
  \centering
  \resizebox{0.9\textwidth}{!}{
  \begin{tabular}{c|ccc|ccc|ccc}
    \toprule
     Dataset & \multicolumn{3}{c|}{CIFAR-10} & \multicolumn{3}{c|}{CIFAR-100} & \multicolumn{3}{c}{Tiny ImageNet} \\
     IPC & 1 & 10 & 50 & 1 & 10 & 50 & 1 & 10 & 50 \\
     Ratio (\%) & 0.02 & 0.2 & 1 & 0.2 & 2 & 10 & 0.2 & 2 & 10 \\
    \midrule
      Random & 14.4±2.0 & 26.0±1.2 & 43.4±1.0 & 4.2±0.3 & 14.6±0.5 & 30.0±0.4 & 1.4±0.1 & 5.0±0.2 & 15.0±0.4\\
     Herding & 21.5±1.2 & 31.6±0.7 & 40.4±0.6 & 8.4±0.3 & 17.3±0.3 & 33.7±0.5 & 2.8±0.2 & 6.3±0.2 & 16.7±0.3\\
     Forgetting & 13.5±1.2 & 23.3±1.0 & 23.3±1.1 & 4.5±0.2 & 15.1±0.3 & 30.5±0.3 & 1.6±0.1 & 5.1±0.2 & 15.0±0.3\\
    \midrule
     % MTT  & 46.2±0.8 & 65.4±0.7 & 71.6±0.2 & 24.3±0.3 & 39.7±0.4 & 47.7±0.2 & 8.8±0.3 & 23.2±0.2 & 28.0±0.3\\
     MTT  & 45.8±0.3 & 64.7±0.5 & 71.5±0.5 & 23.9±1.0 & 38.7±0.4 & 47.3±0.1 & 8.3±0.4 & 20.6±0.2 & 28.0±0.3\\
     % 46.2±0.9 20.7±0.2  47.3±0.1 28.0±0.3
     % \textbf{{+\mymethod}}   & 46.2±0.7& 66.0±0.3& 72.0±0.5 & 24.6±0.3  & 39.8±0.3 & 47.8±0.2 & 8.9±0.2 & xxx & xxx\\
     \textbf{{+\mymethod}}   & 46.2±0.7 & 65.3±0.3& 71.8±0.5 & 24.3±0.3  & 38.8±0.3 & 47.3±0.2 & 8.5±0.2 & 20.7±0.2 &28.0±0.2 \\
     % 46.2±0.7 47.3±0.2 28.0±0.2 \underline{20.7±0.2}
    \midrule
     FTD & 46.7±0.7 & 65.2±0.5 & 72.2±0.1 & 25.1±0.4 & 42.5±0.1 & 50.3±0.3 & {10.9±0.1} & {21.8±0.3} & -\\
     % 42.5±0.1
     \textbf{{+\mymethod}}  & 47.2±0.7& 66.4±0.4 & 73.3±0.4 & 25.4±0.3& 42.6±0.1 & 50.5±0.3 & {11.2±0.1} & {22.2±0.2} & -\\
     % 42.5±0.4
    \midrule
     % TESLA & 48.5±0.8 & 66.4±0.8 & 72.6±0.7 & 24.8±0.4 & 41.7±0.3 & 47.9±0.3 & - & - & -\
     TESLA & 47.4±0.3 & 65.0±0.7 & 71.4±0.5 & 23.9±0.3 & 35.8±0.7 & 44.9±0.4 & - & - & -\\
     % 35.8±0.7
     % \textbf{{+\mymethod}}  & 49.3±0.7 & 66.8±0.4 & 73.1±0.2 & 25.0±0.3 & 41.7±0.2 & 48.0±0.3 & - & - & -\\
     \textbf{{+\mymethod}}  & 47.9±0.7 & 65.3±0.4 & 71.8±0.2 & 24.2±0.2 & 35.9±0.2 & 45.0±0.4 & - & - & -\\
     % 35.8±0.2
    \midrule
     % DATM& 46.9±0.5 & 66.8±0.2 & 76.1±0.3 & 27.9±0.2 & 47.2±0.4 & 55.0±0.2 & 17.1±0.3 & 31.1±0.3 & 39.7±0.3\\
     DATM& 46.1±0.5 & 66.4±0.6 & 75.9±0.3 & 27.7±0.3 & {47.6±0.2} &  {52.1±0.1} & 17.1±0.3 & 30.1±0.3 & 39.7±0.1\\
     % 47.6±0.3 54.1±0.3
     % \textbf{{+\mymethod}} &47.2±0.4&67.0±0.5&76.3±0.2&28.2±0.2 &47.4±0.4 &55.0±0.3 &17.4±0.2 &31.5±0.2 &39.9±0.2\\
     \textbf{{+\mymethod}} &46.4±0.4&66.6±0.4&76.1±0.2&28.0±0.2 &{47.8±0.2}&{52.5±0.2} &17.4±0.2 &30.7±0.2 &39.9±0.2\\
     %\underline{47.6±0.3}  \underline{54.1±0.3} 
    \midrule
    Whole Dataset & \multicolumn{3}{c|}{84.8±0.1} & \multicolumn{3}{c|}{56.2±0.3} & \multicolumn{3}{c}{37.6±0.4} \\
    \bottomrule
  \end{tabular}
  }
  % \vspace{-8pt}
\end{table*}

\textbf{Generalization performance to other architectures.}
We evaluated the generalizability of synthetic datasets generated through distillation. We used DSAC and DATM, which are current SOTA methods in GM-based and TM-based distillation, respectively. After distillation, the synthetic datasets were assessed using various neural networks, including ResNet-18 \cite{he2016deep}, VGG-11 \cite{simonyan2014very}, AlexNet \cite{krizhevsky2012imagenet}, LeNet \cite{lecun1998gradient} and MLP. As shown in  Table~\ref{table:cross}, even though our synthetic datasets were distilled using ConvNet, it generalizes well across most networks. Notably, for the experiment of DATM on CIFAR-10 with IPC = 1, employing SDC resulted in an accuracy improvement of \textbf{4.61\%} when using AlexNet. Employing SDC to DSAC led to an accuracy improvement of \textbf{0.9\%} on SVHN with IPC = 10 when using MLP. Additional results can be found in Appendix~\ref{supp:cross}.

\begin{table}[tb!]
  \caption{Cross-architecture evaluation was conducted on the distilled dataset with (a) IPC = 1 for the TM-based method (DATM) and (b) IPC = 10 for the GM-based method (DSAC), both w/ and w/o {\mymethod}. Adding {\mymethod} improves performance on unseen networks compared to current SOTA methods.}
  % \vspace{-5pt}
  \label{table:cross}
  \centering
  
  \hspace{-1cm}(a)\hspace{0.1cm}% Adjust spacing as needed
  \vspace{8pt}
  \resizebox{0.45\textwidth}{!}{
    \begin{tabular}{cc|ccccc}
    \toprule
     Dataset & Method & ResNet-18 & VGG-11 & AlexNet & LeNet & MLP \\
     \midrule
      \multirow{2}{*}{CIFAR-10} & DATM & 29.62 & 25.12 & 19.38 & 23.41 & \textbf{23.08} \\
      & \textbf{+SDC} & \textbf{31.29} & \textbf{25.99} & \textbf{23.99} & \textbf{23.65} & 22.90 \\
      % \midrule
      \multirow{2}{*}{CIFAR-100} & DATM & 11.52 & 8.74 & 1.95 & 6.71 & 6.47 \\
      & \textbf{+SDC} & \textbf{12.10} & \textbf{8.78} & \textbf{3.73} & \textbf{6.84} & \textbf{6.51} \\
      % \midrule
      \multirow{2}{*}{Tiny ImageNet} & DATM & 4.36 & 5.93 & 4.33 & 2.42 & 2.29 \\
      & \textbf{+SDC} & \textbf{4.74} & \textbf{6.45} & \textbf{4.34} & \textbf{2.79} & \textbf{2.32} \\
    \bottomrule
  \end{tabular}
  }

\bigskip % Optional space between tables

  \hspace{-1cm}(b)\hspace{0.1cm}% Adjust spacing as needed
    \resizebox{0.45\textwidth}{!}{
  \begin{tabular}{cc|ccccc}
    \toprule
     Dataset & Method & ResNet-18 & VGG-11 & AlexNet & LeNet & MLP \\
     \midrule
      \multirow{2}{*}{MNIST} & DSAC & 97.44& 96.88& 95.30& 95.31& 90.62\\
      & \textbf{{+\mymethod}} &\textbf{97.65}&\textbf{97.13}& \textbf{95.73}& \textbf{95.67}&\textbf{90.93}\\
     % \midrule
      \multirow{2}{*}{FashionMNIST} & DSAC & 82.17& 82.59& 80.73& 79.82& 80.09\\
      & \textbf{{+\mymethod}} &\textbf{82.87}& \textbf{82.73}& \textbf{81.15}& \textbf{79.96}& \textbf{80.36}\\
     % \midrule
      \multirow{2}{*}{SVHN} & DSAC & 70.59& 76.44& 49.66& 55.98& 39.11\\
      & \textbf{{+\mymethod}} &\textbf{71.03}& \textbf{76.63}& \textbf{49.78}& \textbf{56.27}& \textbf{40.03}\\
    \bottomrule
  \end{tabular}
  }
% \vspace{-10pt}
\end{table}

\subsection{Further Discussions}\label{sec:discussion}
\begin{figure}[htbp]
    \centering
    \begin{minipage}[t]{0.45\textwidth}
        \centering
        \includegraphics[width=0.8\textwidth]{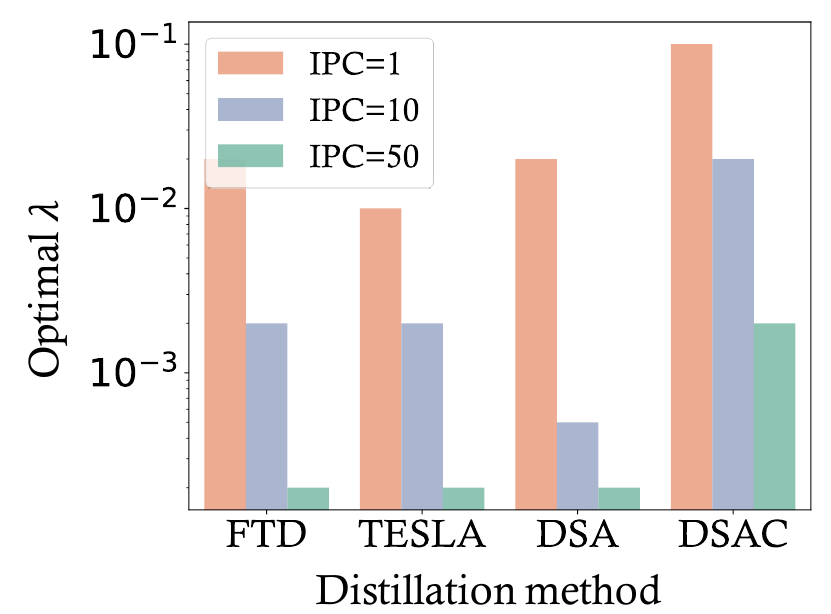}
        % \vspace{5pt}
        \caption{Optimal $\lambda$ values for various matching-based distillation methods with SDC, performed on datasets with different IPC values.}
        \label{fig:different_lambda}
        % \vspace{10pt}
    \end{minipage}
    \hspace{0.02\textwidth}
    \begin{minipage}[t]{0.49\textwidth}
        \centering
        \includegraphics[width=0.9\textwidth]{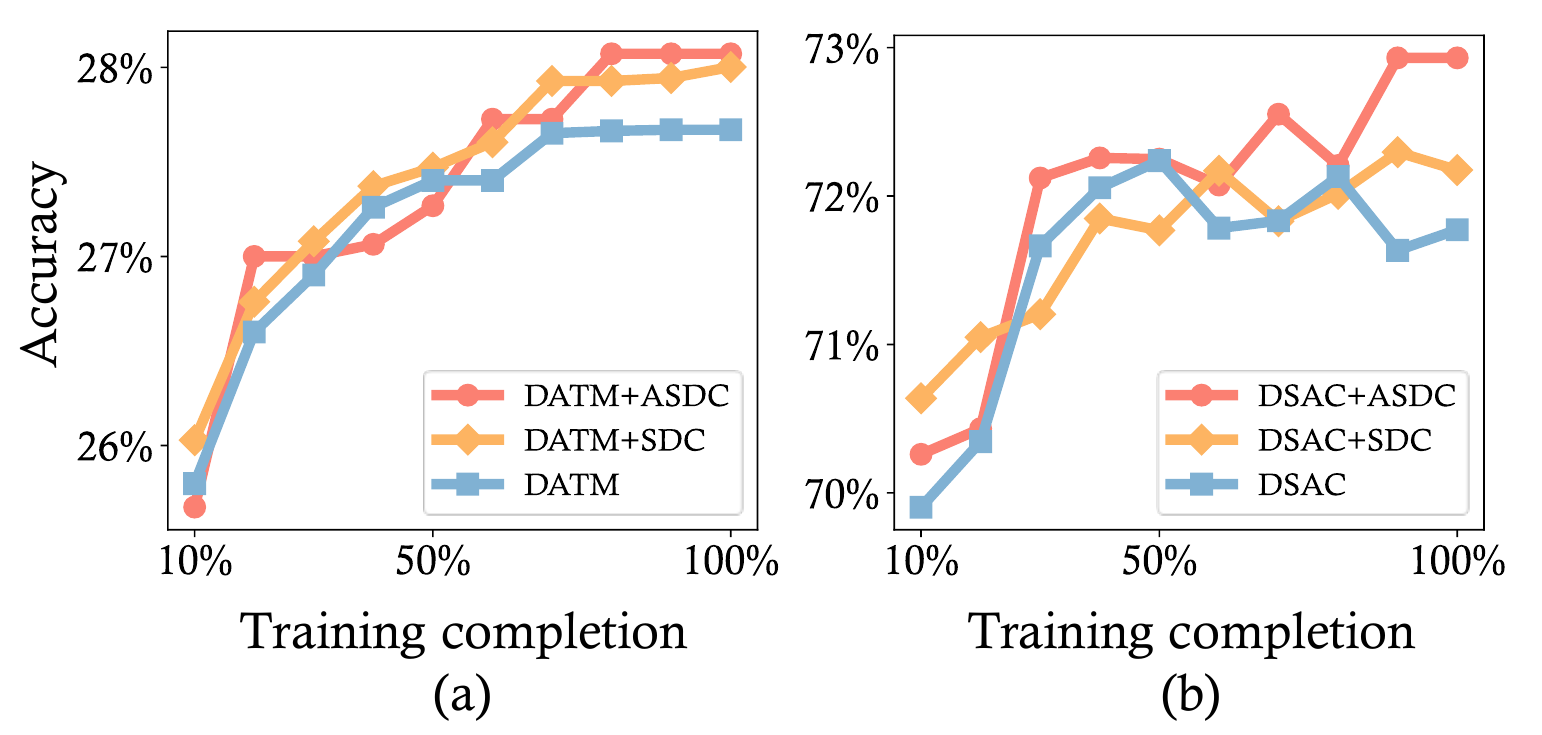}
        \vspace{-5pt}
        \caption{Flexibly adjusting sample difficulty correction with an adaptive increase in $\lambda$ results in higher accuracy compared to standard {\mymethod} and baseline methods. We present the results of (a) DATM and (b) DSAC.}
        \label{fig:increasing_lambda}
        \vspace{-20pt}
    \end{minipage}
\end{figure}

\textbf{Discussion of {\mymethod} coefficient $\lambda$.} 
The selection of the regularization coefficient $\lambda$ is pivotal for the quality of the distilled dataset. Our theory suggests that a larger $\lambda$ typically produces better synthetic datasets for smaller IPC values. Ideally, for low IPC settings, it is better to employ a large $\lambda$ to strongly penalize sample difficulty, whereas, for high IPC settings, the required $\lambda$ can be small or even close to zero in extreme cases. For simplicity and to maintain consistency across different datasets and baseline methods, we have set $\lambda = 0.002$ as the default value in most of our experiments. As demonstrated in Figure~\ref{fig:different_lambda}, this choice of $\lambda$ aligns with the IPC values. Results for FTD and TESLA are based on CIFAR-10, results for DSA are based on SVHN, and results for DSAC are based on MNIST. Additionally, we further show that the choice of $\lambda$ is not sensitive in Appendix~\ref{supp:sensitivity}.

\textbf{Adaptive sample difficulty correction by adaptively increasing $\lambda$ during distillation.} 
While our SDC seeks simplicity in regularization, DATM \cite{guo2023towards} claims that the matching difficulty is increased through optimization. Inspired by their observation, we implemented a strategy where $\lambda$ increases progressively throughout the matching phases. This method is designed to incrementally adjust the focus from easier to more complex patterns. Inspired by their observation, we applied an \textit{Adaptive Sample Difficulty Correction} (ASDC) strategy in our experiments with a TM-based method on the CIFAR-100 with IPC = 1 and with a GM-based method on the FashionMNIST with IPC = 1. The $\lambda$ of DATM was initialized to 0.02 and logarithmically increased to 0.08 over 10,000 iterations and DSAC was initialized to 0.002 and logarithmically increased to 0.008 over 10,000 steps. For DATM, we use max test accuracy, while for DSAC, we use test accuracy. Experimental results of ASDC validate its potential to significantly enhance learning by finetuning regularization according to the complexity of the learned patterns. Figure~\ref{fig:increasing_lambda} illustrates that ASDC further improves our method within SOTA matching methods. Additional results are provided in Appendix~\ref{supp:asdc}.

\section{Conclusion}
In this study, we empirically examine the matching-based dataset distillation method in relation to \textit{sample difficulty}, observing clear trends as measured by gradient norm. Additionally, we adapt a neural scaling law from data pruning to theoretically explain dataset distillation. Our theoretical analysis suggests that for small synthetic datasets, the optimal approach is to generate data using easier samples from the original dataset rather than harder ones. To facilitate this, we propose a simplicity-centric regularization method, termed \emph{Sample Difficulty Correction} (SDC), aimed at improving synthetic data quality by predominantly utilizing easier samples in the data generation process. This method can be easily incorporated to existing matching-based methods, and can be implemented with a few lines of code. Experimental results underscore the importance of proper regularization within the optimization process for dataset distillation. We anticipate that this work will deepen the theoretical understanding of dataset distillation.

{
    \small
    \bibliographystyle{ieeenat_fullname}
    \bibliography{main}

\begin{thebibliography}{52}
\providecommand{\natexlab}[1]{#1}
\providecommand{\url}[1]{\texttt{#1}}
\expandafter\ifx\csname urlstyle\endcsname\relax
  \providecommand{\doi}[1]{doi: #1}\else
  \providecommand{\doi}{doi: \begingroup \urlstyle{rm}\Url}\fi

\bibitem[Abbas et~al.(2023)Abbas, Tirumala, Simig, Ganguli, and Morcos]{abbas2023semdedup}
Amro Abbas, Kushal Tirumala, D{\'a}niel Simig, Surya Ganguli, and Ari~S Morcos.
\newblock Semdedup: Data-efficient learning at web-scale through semantic deduplication.
\newblock \emph{arXiv preprint arXiv:2303.09540}, 2023.

\bibitem[Bohdal et~al.(2020)Bohdal, Yang, and Hospedales]{bohdal2020flexible}
Ondrej Bohdal, Yongxin Yang, and Timothy Hospedales.
\newblock Flexible dataset distillation: Learn labels instead of images.
\newblock \emph{arXiv preprint arXiv:2006.08572}, 2020.

\bibitem[Cazenavette et~al.(2022)Cazenavette, Wang, Torralba, Efros, and Zhu]{cazenavette2022dataset}
George Cazenavette, Tongzhou Wang, Antonio Torralba, Alexei~A Efros, and Jun-Yan Zhu.
\newblock Dataset distillation by matching training trajectories.
\newblock In \emph{Proceedings of the IEEE/CVF Conference on Computer Vision and Pattern Recognition}, pages 4750--4759, 2022.

\bibitem[Chen et~al.(2022)Chen, Kerkouche, and Fritz]{chen2022private}
Dingfan Chen, Raouf Kerkouche, and Mario Fritz.
\newblock Private set generation with discriminative information.
\newblock \emph{Advances in Neural Information Processing Systems}, 35:\penalty0 14678--14690, 2022.

\bibitem[Choi et~al.(2021)Choi, Yi, Kim, Choo, Kim, Chang, Gwon, and Chang]{choi2021vab}
Jongwon Choi, Kwang~Moo Yi, Jihoon Kim, Jinho Choo, Byoungjip Kim, Jinyeop Chang, Youngjune Gwon, and Hyung~Jin Chang.
\newblock Vab-al: Incorporating class imbalance and difficulty with variational bayes for active learning.
\newblock In \emph{Proceedings of the IEEE/CVF conference on computer vision and pattern recognition}, pages 6749--6758, 2021.

\bibitem[Chung et~al.(2023)Chung, Chou, Yu, Chen, Kuo, and Ho]{chung2023rethinking}
Ming-Yu Chung, Sheng-Yen Chou, Chia-Mu Yu, Pin-Yu Chen, Sy-Yen Kuo, and Tsung-Yi Ho.
\newblock Rethinking backdoor attacks on dataset distillation: A kernel method perspective.
\newblock \emph{arXiv preprint arXiv:2311.16646}, 2023.

\bibitem[Coleman et~al.(2019)Coleman, Yeh, Mussmann, Mirzasoleiman, Bailis, Liang, Leskovec, and Zaharia]{coleman2019selection}
Cody Coleman, Christopher Yeh, Stephen Mussmann, Baharan Mirzasoleiman, Peter Bailis, Percy Liang, Jure Leskovec, and Matei Zaharia.
\newblock Selection via proxy: Efficient data selection for deep learning.
\newblock \emph{arXiv preprint arXiv:1906.11829}, 2019.

\bibitem[Cui et~al.(2023)Cui, Wang, Si, and Hsieh]{cui2023scaling}
Justin Cui, Ruochen Wang, Si Si, and Cho-Jui Hsieh.
\newblock Scaling up dataset distillation to imagenet-1k with constant memory.
\newblock In \emph{International Conference on Machine Learning}, pages 6565--6590. PMLR, 2023.

\bibitem[Cui et~al.(2024)Cui, Zhang, Deng, Dong, and Zhu]{cui2024learning}
Peng Cui, Dan Zhang, Zhijie Deng, Yinpeng Dong, and Jun Zhu.
\newblock Learning sample difficulty from pre-trained models for reliable prediction.
\newblock \emph{Advances in Neural Information Processing Systems}, 36, 2024.

\bibitem[Deng(2012)]{deng2012mnist}
Li Deng.
\newblock The mnist database of handwritten digit images for machine learning research.
\newblock \emph{IEEE Signal Processing Magazine}, 29\penalty0 (6):\penalty0 141--142, 2012.

\bibitem[Dong et~al.(2022)Dong, Zhao, and Lyu]{dong2022privacy}
Tian Dong, Bo Zhao, and Lingjuan Lyu.
\newblock Privacy for free: How does dataset condensation help privacy?
\newblock In \emph{International Conference on Machine Learning}, pages 5378--5396. PMLR, 2022.

\bibitem[Du et~al.(2023)Du, Jiang, Tan, Zhou, and Li]{du2023minimizing}
Jiawei Du, Yidi Jiang, Vincent~YF Tan, Joey~Tianyi Zhou, and Haizhou Li.
\newblock Minimizing the accumulated trajectory error to improve dataset distillation.
\newblock In \emph{Proceedings of the IEEE/CVF Conference on Computer Vision and Pattern Recognition}, pages 3749--3758, 2023.

\bibitem[Ghorbani and Zou(2019)]{ghorbani2019data}
Amirata Ghorbani and James Zou.
\newblock Data shapley: Equitable valuation of data for machine learning.
\newblock In \emph{International conference on machine learning}, pages 2242--2251. PMLR, 2019.

\bibitem[Gu et~al.(2023)Gu, Wang, Jiang, and You]{gu2023summarizing}
Jianyang Gu, Kai Wang, Wei Jiang, and Yang You.
\newblock Summarizing stream data for memory-restricted online continual learning.
\newblock \emph{arXiv preprint arXiv:2305.16645}, 2023.

\bibitem[Guo et~al.(2023)Guo, Wang, Cazenavette, Li, Zhang, and You]{guo2023towards}
Ziyao Guo, Kai Wang, George Cazenavette, Hui Li, Kaipeng Zhang, and Yang You.
\newblock Towards lossless dataset distillation via difficulty-aligned trajectory matching.
\newblock \emph{arXiv preprint arXiv:2310.05773}, 2023.

\bibitem[He et~al.(2016)He, Zhang, Ren, and Sun]{he2016deep}
Kaiming He, Xiangyu Zhang, Shaoqing Ren, and Jian Sun.
\newblock Deep residual learning for image recognition.
\newblock In \emph{Proceedings of the IEEE conference on computer vision and pattern recognition}, pages 770--778, 2016.

\bibitem[Kaplan et~al.(2020)Kaplan, McCandlish, Henighan, Brown, Chess, Child, Gray, Radford, Wu, and Amodei]{kaplan2020scaling}
Jared Kaplan, Sam McCandlish, Tom Henighan, Tom~B Brown, Benjamin Chess, Rewon Child, Scott Gray, Alec Radford, Jeffrey Wu, and Dario Amodei.
\newblock Scaling laws for neural language models.
\newblock \emph{arXiv preprint arXiv:2001.08361}, 2020.

\bibitem[Kim et~al.(2022)Kim, Kim, Oh, Yun, Song, Jeong, Ha, and Song]{kim2022dataset}
Jang-Hyun Kim, Jinuk Kim, Seong~Joon Oh, Sangdoo Yun, Hwanjun Song, Joonhyun Jeong, Jung-Woo Ha, and Hyun~Oh Song.
\newblock Dataset condensation via efficient synthetic-data parameterization.
\newblock In \emph{International Conference on Machine Learning}, pages 11102--11118. PMLR, 2022.

\bibitem[Krizhevsky et~al.(2009)Krizhevsky, Hinton, et~al.]{krizhevsky2009learning}
Alex Krizhevsky, Geoffrey Hinton, et~al.
\newblock Learning multiple layers of features from tiny images.
\newblock 2009.

\bibitem[Krizhevsky et~al.(2012)Krizhevsky, Sutskever, and Hinton]{krizhevsky2012imagenet}
Alex Krizhevsky, Ilya Sutskever, and Geoffrey~E Hinton.
\newblock Imagenet classification with deep convolutional neural networks.
\newblock \emph{Advances in neural information processing systems}, 25, 2012.

\bibitem[Le and Yang(2015)]{le2015tiny}
Ya Le and Xuan Yang.
\newblock Tiny imagenet visual recognition challenge.
\newblock \emph{CS 231N}, 7\penalty0 (7):\penalty0 3, 2015.

\bibitem[LeCun et~al.(1998)LeCun, Bottou, Bengio, and Haffner]{lecun1998gradient}
Yann LeCun, L{\'e}on Bottou, Yoshua Bengio, and Patrick Haffner.
\newblock Gradient-based learning applied to document recognition.
\newblock \emph{Proceedings of the IEEE}, 86\penalty0 (11):\penalty0 2278--2324, 1998.

\bibitem[Lee et~al.(2022)Lee, Chun, Jung, Yun, and Yoon]{lee2022dataset}
Saehyung Lee, Sanghyuk Chun, Sangwon Jung, Sangdoo Yun, and Sungroh Yoon.
\newblock Dataset condensation with contrastive signals.
\newblock In \emph{International Conference on Machine Learning}, pages 12352--12364. PMLR, 2022.

\bibitem[Li et~al.(2024)Li, Chen, Chen, He, Gu, and Zhou]{li2024selective}
Ming Li, Lichang Chen, Jiuhai Chen, Shwai He, Jiuxiang Gu, and Tianyi Zhou.
\newblock Selective reflection-tuning: Student-selected data recycling for llm instruction-tuning.
\newblock \emph{arXiv preprint arXiv:2402.10110}, 2024.

\bibitem[Lin et~al.(2024)Lin, Wang, Li, Yang, Feng, Wei, and Chua]{lin2024data}
Xinyu Lin, Wenjie Wang, Yongqi Li, Shuo Yang, Fuli Feng, Yinwei Wei, and Tat-Seng Chua.
\newblock Data-efficient fine-tuning for llm-based recommendation.
\newblock \emph{arXiv preprint arXiv:2401.17197}, 2024.

\bibitem[Loo et~al.(2023)Loo, Hasani, Lechner, Amini, and Rus]{loo2023understanding}
Noel Loo, Ramin Hasani, Mathias Lechner, Alexander Amini, and Daniela Rus.
\newblock Understanding reconstruction attacks with the neural tangent kernel and dataset distillation.
\newblock \emph{arXiv preprint arXiv:2302.01428}, 2023.

\bibitem[Maharana et~al.(2023)Maharana, Yadav, and Bansal]{maharana2023d2}
Adyasha Maharana, Prateek Yadav, and Mohit Bansal.
\newblock D2 pruning: Message passing for balancing diversity and difficulty in data pruning.
\newblock \emph{arXiv preprint arXiv:2310.07931}, 2023.

\bibitem[Maroto and Frossard(2024)]{maroto2024puma}
Javier Maroto and Pascal Frossard.
\newblock Puma: margin-based data pruning.
\newblock \emph{arXiv preprint arXiv:2405.06298}, 2024.

\bibitem[Masarczyk and Tautkute(2020)]{masarczyk2020reducing}
Wojciech Masarczyk and Ivona Tautkute.
\newblock Reducing catastrophic forgetting with learning on synthetic data.
\newblock In \emph{Proceedings of the IEEE/CVF Conference on Computer Vision and Pattern Recognition Workshops}, pages 252--253, 2020.

\bibitem[Meding et~al.(2022)Meding, Buschoff, Geirhos, and Wichmann]{meding2022trivial}
Kristof Meding, Luca M.~Schulze Buschoff, Robert Geirhos, and Felix~A. Wichmann.
\newblock Trivial or impossible -- dichotomous data difficulty masks model differences (on imagenet and beyond), 2022.

\bibitem[Medvedev and D’yakonov(2021)]{medvedev2021learning}
Dmitry Medvedev and Alexander D’yakonov.
\newblock Learning to generate synthetic training data using gradient matching and implicit differentiation.
\newblock In \emph{International Conference on Analysis of Images, Social Networks and Texts}, pages 138--150. Springer, 2021.

\bibitem[M{\'e}zard et~al.(1987)M{\'e}zard, Parisi, and Virasoro]{mezard1987spin}
Marc M{\'e}zard, Giorgio Parisi, and Miguel~Angel Virasoro.
\newblock \emph{Spin glass theory and beyond: An Introduction to the Replica Method and Its Applications}.
\newblock World Scientific Publishing Company, 1987.

\bibitem[Netzer et~al.(2011)Netzer, Wang, Coates, Bissacco, Wu, Ng, et~al.]{netzer2011reading}
Yuval Netzer, Tao Wang, Adam Coates, Alessandro Bissacco, Baolin Wu, Andrew~Y Ng, et~al.
\newblock Reading digits in natural images with unsupervised feature learning.
\newblock In \emph{NIPS workshop on deep learning and unsupervised feature learning}, page~7. Granada, Spain, 2011.

\bibitem[Paul et~al.(2023)Paul, Ganguli, and Dziugaite]{paul2023deep}
Mansheej Paul, Surya Ganguli, and Gintare~Karolina Dziugaite.
\newblock Deep learning on a data diet: Finding important examples early in training, 2023.

\bibitem[Rosasco et~al.(2021)Rosasco, Carta, Cossu, Lomonaco, and Bacciu]{rosasco2021distilled}
Andrea Rosasco, Antonio Carta, Andrea Cossu, Vincenzo Lomonaco, and Davide Bacciu.
\newblock Distilled replay: Overcoming forgetting through synthetic samples, 2021.

\bibitem[Sachdeva and McAuley(2023)]{sachdeva2023data}
Noveen Sachdeva and Julian McAuley.
\newblock Data distillation: A survey.
\newblock \emph{arXiv preprint arXiv:2301.04272}, 2023.

\bibitem[Simonyan and Zisserman(2014)]{simonyan2014very}
Karen Simonyan and Andrew Zisserman.
\newblock Very deep convolutional networks for large-scale image recognition.
\newblock \emph{arXiv preprint arXiv:1409.1556}, 2014.

\bibitem[Sorscher et~al.(2022)Sorscher, Geirhos, Shekhar, Ganguli, and Morcos]{sorscher2022beyond}
Ben Sorscher, Robert Geirhos, Shashank Shekhar, Surya Ganguli, and Ari Morcos.
\newblock Beyond neural scaling laws: beating power law scaling via data pruning.
\newblock \emph{Advances in Neural Information Processing Systems}, 35:\penalty0 19523--19536, 2022.

\bibitem[Such et~al.(2020)Such, Rawal, Lehman, Stanley, and Clune]{such2020generative}
Felipe~Petroski Such, Aditya Rawal, Joel Lehman, Kenneth Stanley, and Jeffrey Clune.
\newblock Generative teaching networks: Accelerating neural architecture search by learning to generate synthetic training data.
\newblock In \emph{International Conference on Machine Learning}, pages 9206--9216. PMLR, 2020.

\bibitem[Tan et~al.(2024)Tan, Wu, Du, Chen, Wang, Wang, and Qi]{tan2024data}
Haoru Tan, Sitong Wu, Fei Du, Yukang Chen, Zhibin Wang, Fan Wang, and Xiaojuan Qi.
\newblock Data pruning via moving-one-sample-out.
\newblock \emph{Advances in Neural Information Processing Systems}, 36, 2024.

\bibitem[Toneva et~al.(2018)Toneva, Sordoni, Combes, Trischler, Bengio, and Gordon]{toneva2018empirical}
Mariya Toneva, Alessandro Sordoni, Remi Tachet~des Combes, Adam Trischler, Yoshua Bengio, and Geoffrey~J Gordon.
\newblock An empirical study of example forgetting during deep neural network learning.
\newblock \emph{arXiv preprint arXiv:1812.05159}, 2018.

\bibitem[Wang et~al.(2025{\natexlab{a}})Wang, Jin, Wang, Wang, Zhang, Li, Wen, Li, He, Hu, and Zhang]{wang2025datawhisperer}
Shaobo Wang, Xiangqi Jin, Ziming Wang, Jize Wang, Jiajun Zhang, Kaixin Li, Zichen Wen, Zhong Li, Conghui He, Xuming Hu, and Linfeng Zhang.
\newblock Data whisperer: Efficient data selection for task-specific llm fine-tuning via few-shot in-context learning.
\newblock \emph{Annual Meeting of the Association for Computational Linguistics}, 2025{\natexlab{a}}.

\bibitem[Wang et~al.(2025{\natexlab{b}})Wang, Yang, Liu, Sun, Hu, He, and Zhang]{NCFM}
Shaobo Wang, Yicun Yang, Zhiyuan Liu, Chenghao Sun, Xuming Hu, Conghui He, and Linfeng Zhang.
\newblock Dataset distillation with neural characteristic function: A minmax perspective.
\newblock In \emph{Proceedings of the IEEE conference on computer vision and pattern recognition}, 2025{\natexlab{b}}.

\bibitem[Wang et~al.(2025{\natexlab{c}})Wang, Yang, Zhang, Sun, Li, Hu, and Zhang]{DRUPI}
Shaobo Wang, Yantai Yang, Shuaiyu Zhang, Chenghao Sun, Weiya Li, Xuming Hu, and Linfeng Zhang.
\newblock {DRUPI}: Dataset reduction using privileged information.
\newblock In \emph{The Future of Machine Learning Data Practices and Repositories at ICLR 2025}, 2025{\natexlab{c}}.

\bibitem[Wang et~al.(2018)Wang, Zhu, Torralba, and Efros]{wang2018dataset}
Tongzhou Wang, Jun-Yan Zhu, Antonio Torralba, and Alexei~A Efros.
\newblock Dataset distillation.
\newblock \emph{arXiv preprint arXiv:1811.10959}, 2018.

\bibitem[Welling(2009)]{welling2009herding}
Max Welling.
\newblock Herding dynamical weights to learn.
\newblock In \emph{Proceedings of the 26th annual international conference on machine learning}, pages 1121--1128, 2009.

\bibitem[Xiao et~al.(2017)Xiao, Rasul, and Vollgraf]{xiao2017fashionmnist}
Han Xiao, Kashif Rasul, and Roland Vollgraf.
\newblock Fashion-mnist: a novel image dataset for benchmarking machine learning algorithms, 2017.

\bibitem[Xie et~al.(2020)Xie, Feng, Chen, Sun, Ma, and Song]{xie2020deal}
Shuai Xie, Zunlei Feng, Ying Chen, Songtao Sun, Chao Ma, and Mingli Song.
\newblock Deal: Difficulty-aware active learning for semantic segmentation.
\newblock In \emph{Proceedings of the Asian conference on computer vision}, 2020.

\bibitem[Xu* et~al.(2025)Xu*, Wang*, Sun, Zhang, and Zhang]{xu2025unseen}
Furui Xu*, Shaobo Wang*, Chenghao Sun, Jiajun Zhang, and Linfeng Zhang.
\newblock Rethink dataset pruning from a generalization perspective.
\newblock \emph{The Future of Machine Learning Data Practices and Repositories at ICLR 2025}, 2025.

\bibitem[Yang et~al.(2023)Yang, Shen, Wang, Liu, and Guo]{yang2023efficient}
Enneng Yang, Li Shen, Zhenyi Wang, Tongliang Liu, and Guibing Guo.
\newblock An efficient dataset condensation plugin and its application to continual learning.
\newblock \emph{Advances in Neural Information Processing Systems}, 36, 2023.

\bibitem[Zhao and Bilen(2021)]{zhao2021dataset}
Bo Zhao and Hakan Bilen.
\newblock Dataset condensation with differentiable siamese augmentation.
\newblock In \emph{International Conference on Machine Learning}, pages 12674--12685. PMLR, 2021.

\bibitem[Zhao et~al.(2020)Zhao, Mopuri, and Bilen]{zhao2020dataset}
Bo Zhao, Konda~Reddy Mopuri, and Hakan Bilen.
\newblock Dataset condensation with gradient matching.
\newblock \emph{arXiv preprint arXiv:2006.05929}, 2020.

\end{thebibliography}
}

\clearpage
\setcounter{page}{1}
\setcounter{section}{0}
\maketitlesupplementary

%%%%%%%%%%%%%%%%%%%%%%%%%%%%%%%%%%%%%%%%%%%%%%%%%%%%%%%%%%%%

\section{More Result on the Relationships between Sample Difficulty, Gardient Norm, and Loss}\label{supp:difficulty}

\begin{figure*}[tb!]
    \centering
    \includegraphics[width=0.99\textwidth]{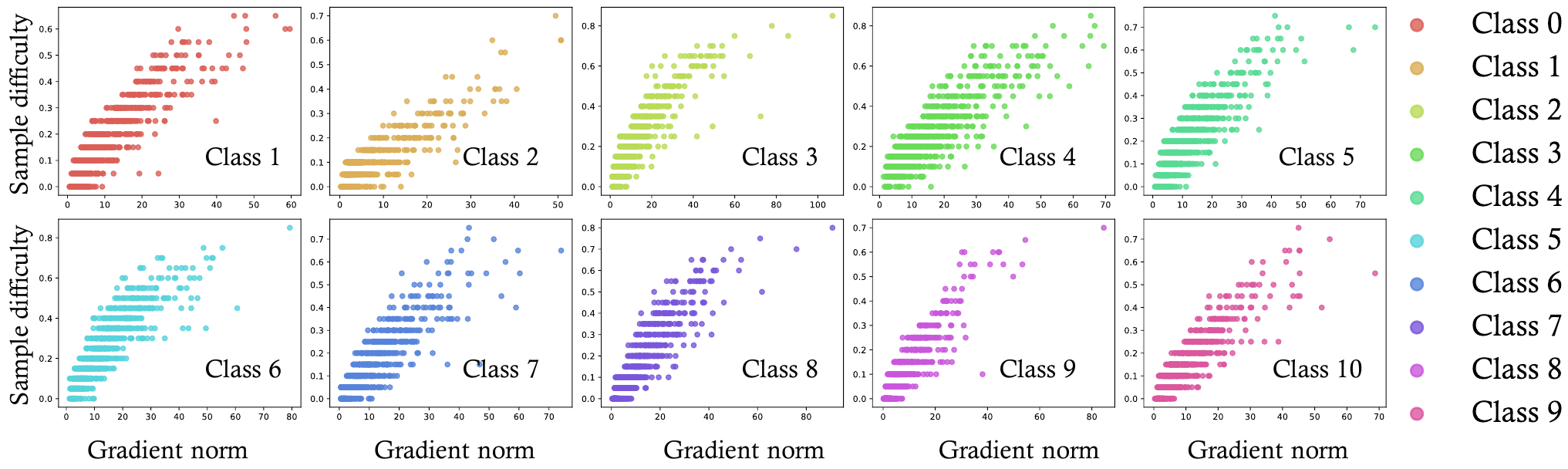}
    \caption{The statistical relationship between sample difficulty $\chi(x,y;\Theta_t)$, gradient norm $\text{GraDN}(x,y;\Theta_t)$ for each sample $(x,y)$ on a series of ResNet-18 models with parameters $\theta_t\in \Theta_t$ on CIFAR-10.  1000 samples were randomly selected for each category.}
    \label{fig_supp:cifar10_grad}
\end{figure*}

\begin{figure*}[tb!]
    \centering
    \includegraphics[width=0.99\textwidth]{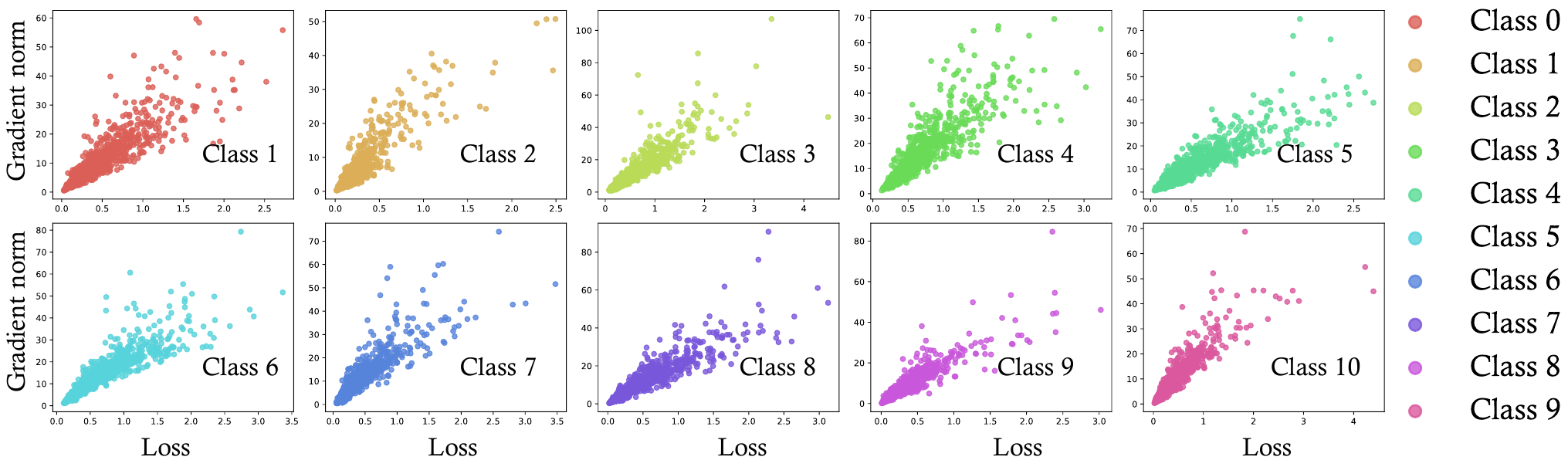}
    \caption{The statistical relationship between gradient norm $\text{GraDN}(x,y;\Theta_t)$  and average validation loss for each sample $(x,y)$ on a series of ResNet-18 models with parameters $\theta_t\in \Theta_t$ on CIFAR-10. 1000 samples were randomly selected for each category.}
    \label{fig_supp:cifar10_grad_loss}
\end{figure*}

\begin{figure*}[tb!]
    \centering
    \includegraphics[width=0.99\textwidth]{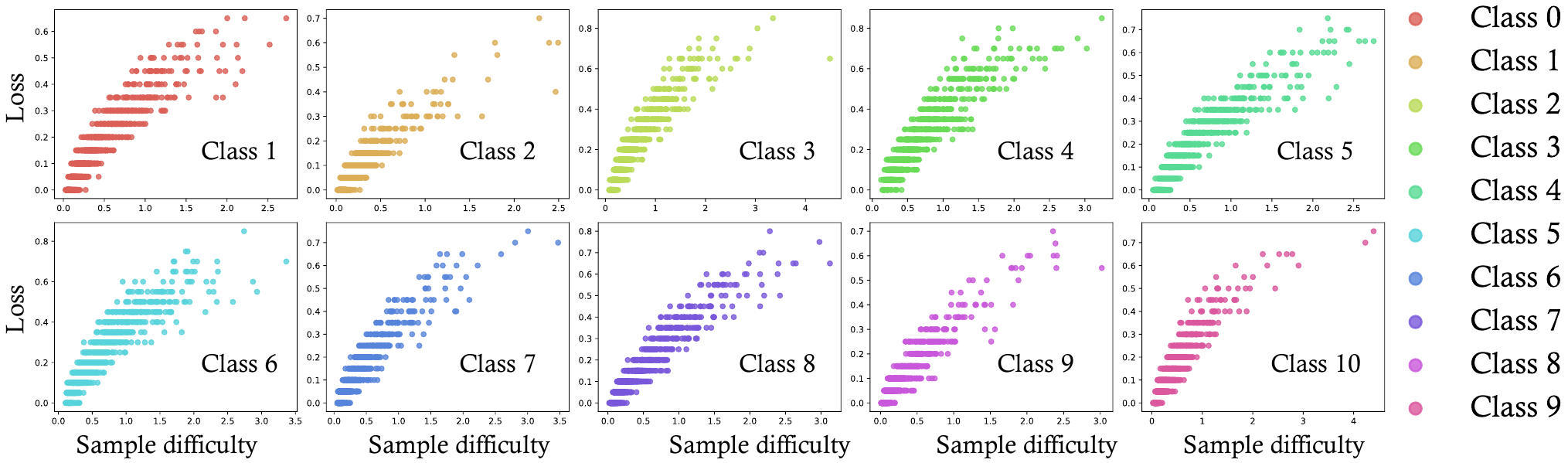}
    \caption{The statistical relationship between sample difficulty $\chi(x,y;\Theta_t)$ and gradient norm $\text{GraDN}(x,y;\Theta_t)$ for each sample $(x,y)$ on a series of ResNet-18 models with parameters $\theta_t\in \Theta_t$ on CIFAR-10. 1000 samples were randomly selected for each category.}
    \label{fig_supp:cifar10_loss}
    \vspace{-10pt}
\end{figure*}

\begin{figure*}[tb!]
    \centering
    \includegraphics[width=0.99\textwidth]{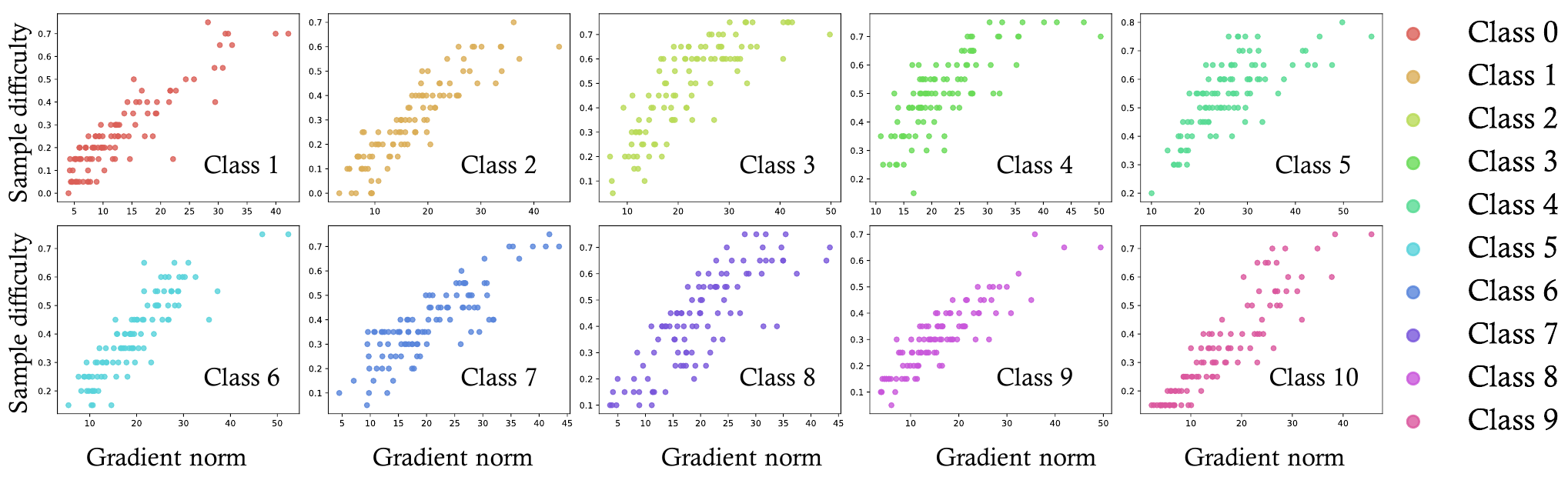}
    \caption{The statistical relationship between sample difficulty $\chi(x,y;\Theta_t)$, gradient norm $\text{GraDN}(x,y;\Theta_t)$ for each sample $(x,y)$ on a series of ResNet-34 models with parameters $\theta_t\in \Theta_t$ on CIFAR-100. 100 samples were randomly selected for each category.}
    \label{fig_supp:cifar100_grad}
    \vspace{-10pt}
\end{figure*}

\begin{figure*}[tb!]
    \centering
    \includegraphics[width=0.99\textwidth]{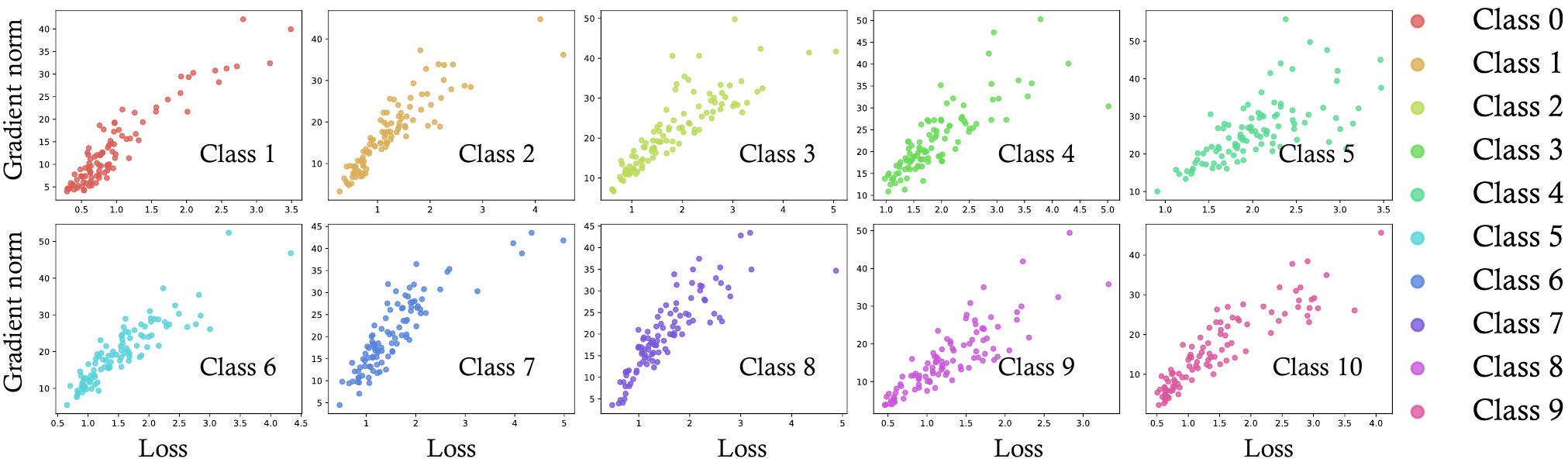}
    \caption{The statistical relationship between gradient norm $\text{GraDN}(x,y;\Theta_t)$  and average validation loss for each sample $(x,y)$ on a series of ResNet-34 models with parameters $\theta_t\in \Theta_t$ on CIFAR-100. 100 samples were randomly selected for each category.}
    \vspace{-10pt}
    \label{fig_supp:cifar100_grad_loss}
\end{figure*}

\begin{figure*}[tb!]
    \centering
    \includegraphics[width=0.99\textwidth]{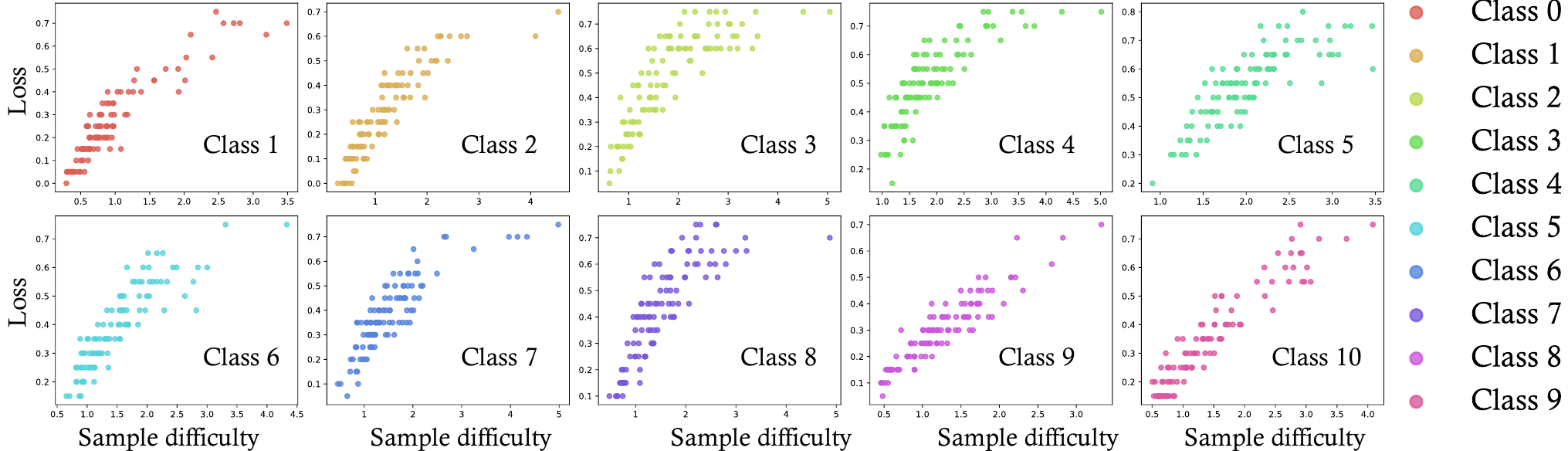}
    \caption{The statistical relationship between sample difficulty $\chi(x,y;\Theta_t)$ and gradient norm $\text{GraDN}(x,y;\Theta_t)$ for each sample $(x,y)$ on a series of ResNet-34 models with parameters $\theta_t\in \Theta_t$ on CIFAR-100. 100 samples were randomly selected for each category.}
    \vspace{-10pt}
    \label{fig_supp:cifar100_loss}
\end{figure*}

In this section, we present further findings on the relationships between sample difficulty $\chi(x,y;\Theta_t)$, gradient norm $\text{GraDN}(x,y;\Theta_t)$, and the average validation loss for each sample $(x,y)$ across a range of models characterized by $\theta_t \in \Theta_t$. The experiments were conducted using ResNet-18 on CIFAR-10 and ResNet-34 on CIFAR-100. Each metric was evaluated across 20 pretrained models. We randomly selected 1000 samples for each category in CIFAR-10, and 100 samples for each category in CIFAR-100. For CIFAR-100, 10 categories for visualization were randomly selected for visualization purposes. As shown in Figure~\ref{fig_supp:cifar10_grad}, Figure~\ref{fig_supp:cifar10_grad_loss}, Figure~\ref{fig_supp:cifar10_loss}, Figure~\ref{fig_supp:cifar100_grad}, Figure~\ref{fig_supp:cifar100_grad_loss}, Figure~\ref{fig_supp:cifar100_loss}, it reveals a significant positive correlation between sample difficulty, gradient norm and loss.

% \clearpage

\section{More Details of Experiments}
\subsection{Parameter Tables}\label{supp:params}

\subsubsection{GM-based Methods}\label{supp:GM}
Regarding the GM-based methods, Table~\ref{tab:hyperparametersGM} provides the corresponding $\lambda$ values after applying \mymethod. All results are obtained from a single experiment, and evaluated 20 times. Baseline results are obtained using identical configurations with the original methods' implementations (please refer to DC and DSA\footnote{\url{https://github.com/VICO-UoE/DatasetCondensation}}, and DCC\footnote{\url{https://github.com/Saehyung-Lee/DCC}}).
% (training networks是什么) . 
Experiments with our {\mymethod} share consistent hyperparameters with the corresponding baselines.

% 我在重写一遍你看看是不是这个意思

% Baseline results, using the settings described in addition to the specified parameters, are largely consistent with the original methods except the inclusion of $\lambda$. For detailed configurations, please refer to DC and DSA\footnote{\url{https://github.com/VICO-UoE/DatasetCondensation}}, and DCC\footnote{\url{https://github.com/Saehyung-Lee/DCC}}.

% Our baseline results with the settings in addition to the above parameters, other than the original method, are largely consistent (refer to DC and DSA at \url{https://github.com/VICO-UoE/DatasetCondensation}, and DCC at \url{https://github.com/Saehyung-Lee/DCC}), experiments of adding SDC were used and baseline with the same parameters, just on the basis of the use of $\lambda$. We set the batch size of both the real data and training networks to 256. The number of internal and external iterations can be found in the above GitHub link. 
\begin{table*}[htbp]
\centering
\caption{The $\lambda$ used for GM-based methods}
\label{tab:hyperparametersGM}
\resizebox{0.7\textwidth}{!}{
\begin{tabular}{c|ccc|ccc|ccc}
\toprule
Dataset & \multicolumn{3}{c|}{MNIST} & \multicolumn{3}{c|}{FashionMNIST} & \multicolumn{3}{c}{SVHN} \\ \cmidrule(r){2-4} \cmidrule(r){5-7} \cmidrule(r){8-10}
IPC & 1 & 10 & 50 & 1 & 10 & 50 & 1 & 10 & 50 \\ \midrule
DC & 0.001 & 0.0005 & 0.001 & 0.0002 & 0.001 & 0.01 & 0.001 & 0.0005 & 0.0002 \\
DSA & 0.001 & 0.00002 & 0.001 & 0.0002 & 0.002 & 0.001 & 0.005 & 0.0005 & 0.01 \\
DSAC & 0.001 & 0.02 & 0.02 & 0.002 & 0.02 & 0.02 & 0.005 & 0.02 & 0.02 \\ \bottomrule
\end{tabular}
}
% \vspace{-15pt}
\end{table*}

\subsubsection{TM-based Methods}\label{supp:TM}

The hyperparameters used in our TM-based methods differ slightly from the original methods (see original implementations of MTT\footnote{\url{https://github.com/GeorgeCazenavette/mtt-distillation}}, DATM\footnote{\url{https://github.com/NUS-HPC-AI-Lab/DATM}}, TESLA\footnote{\url{https://github.com/justincui03/tesla}}, and FTD\footnote{\url{https://github.com/AngusDujw/FTD-distillation}}), particularly in terms of synthesis steps, number of evaluations, and evaluation interval. Our baseline results used the settings in Table~\ref{tab:hyperparametersMTT}, Table~\ref{tab:hyperparametersTESLA}, Table~\ref{tab:hyperparametersFTD} and Table~\ref{tab:hyperparametersDATM}. The experiments of applying {\mymethod} were conducted in the same setting as in the baselines.
In Table~\ref{tab:hyperparametersTESLA}, and Table~\ref{tab:hyperparametersFTD}, we report the optimal hyperparameters using the ConvNetD3 network. All combinations in Table~\ref{tab:hyperparametersFTD} and Table~\ref{tab:hyperparametersDATM} used the ZCA.

\begin{table*}[htbp]
\centering
\caption{Optimal hyperparameters for MTT. A synthesis batch size of ``-'' means that we used the full support set at each synthesis step. }
\label{tab:hyperparametersMTT}

\resizebox{1.0\textwidth}{!}{
\begin{tabular}{c|c|c|c|c|c|c|c|c|c|c|c|c|c}
\toprule
Dataset & Model & IPC & ZCA & \begin{tabular}[c]{@{}c@{}}Synthetic \\ Steps \\ (N)\end{tabular} & \begin{tabular}[c]{@{}c@{}}Expert \\ Epochs \\ (M\textsuperscript{\dag})\end{tabular} & \begin{tabular}[c]{@{}c@{}}Max Start \\ Epoch \\ (T\textsuperscript{+})\end{tabular} & \begin{tabular}[c]{@{}c@{}}Synthetic \\ Batch \\ Size \end{tabular} & \begin{tabular}[c]{@{}c@{}}Learning \\ Rate \\ (Pixels)\end{tabular} & \begin{tabular}[c]{@{}c@{}}Learning \\ Rate \\ (Step Size)\end{tabular} & \begin{tabular}[c]{@{}c@{}}Starting \\ Synthetic \\ Step Size \end{tabular} & \begin{tabular}[c]{@{}c@{}}Num \\ Eval\end{tabular} & \begin{tabular}[c]{@{}c@{}}Eval \\ Iteration\end{tabular} &  $\lambda$ \\ \midrule
\multirow{3}{*}{CIFAR-10} & \multirow{3}{*}{ConvNetD3} & 1 & Y & 50 & 2 & 2 & - & $10^2$ & $10^{-7}$ & $10^{-2}$ & 5 & 100 & 0.0005 \\
 & & 10 & Y & 30 & 2 & 20 & - & $10^2$ & $10^{-4}$ & $10^{-2}$ & 5 & 100 & 0.02 \\
 & & 50 & N & 30 & 2 & 40 & - & $10^3$ & $10^{-5}$ & $10^{-3}$ & 5 & 100 & 0.0002 \\ \midrule
\multirow{3}{*}{CIFAR-100} & \multirow{3}{*}{ConvNetD3} & 1 & Y & 20 & 3 & 20 & - & $10^3$ & $10^{-5}$ & $10^{-2}$ & 5 & 100 & 0.001 \\
 & & 10 & N & 20 & 2 & 20 & - & $10^3$ & $10^{-5}$ & $10^{-2}$ & 5 & 100 & 0.02 \\
 & & 50 & Y & 80 & 2 & 40 & 125 & $10^3$ & $10^{-5}$ & $10^{-2}$ & 5 & 100 & 0.002 \\ \midrule
\multirow{3}{*}{\begin{tabular}[c]{@{}c@{}}Tiny \\ ImageNet\end{tabular}} & \multirow{3}{*}{ConvNetD4} & 1 & N & 10 & 2 & 10 & - & $10^4$ & $10^{-4}$ & $10^{-2}$ & 5 & 100 & 0.005 \\
 & & 10 & N & 20 & 2 & 40 & 200 & $10^4$ & $10^{-4}$ & $10^{-2}$ & 3 & 200 & 0.02 \\
 & & 50 & N & 20 & 2 & 40 & 300 & $10^4$ & $10^{-4}$ & $10^{-2}$ & 3 & 200 & 0.02 \\ \bottomrule
\end{tabular}
}
\end{table*}

\begin{table*}[htbp]
\centering
\caption{Optimal hyperparameters for TESLA. A synthesis batch size of ``-'' means that we used the full support set at each synthesis step. }
\label{tab:hyperparametersTESLA}

\resizebox{1.0\textwidth}{!}{
\begin{tabular}{c|c|c|c|c|c|c|c|c|c|c}
\toprule
Dataset  & IPC & \begin{tabular}[c]{@{}c@{}}Matching \\ Steps\end{tabular} & \begin{tabular}[c]{@{}c@{}}Teacher \\ Epochs\end{tabular} & \begin{tabular}[c]{@{}c@{}}Max Start \\ Epoch\end{tabular} & \begin{tabular}[c]{@{}c@{}}Synthetic \\Batch \\ Size\end{tabular} & \begin{tabular}[c]{@{}c@{}}Learning Rate\\ (Pixels)\end{tabular} & \begin{tabular}[c]{@{}c@{}}Learning \\ Rate \\ (Step Size)\end{tabular} & \begin{tabular}[c]{@{}c@{}}Starting \\ Synthetic \\ Step Size\end{tabular} & ZCA &  $\lambda$ \\ \midrule
\multirow{3}{*}{CIFAR-10}  & 1 & 50 & 2 & 3 & - & $10^2$ & $10^{-7}$ & $10^{-2}$ & Y & 0.01 \\
 &  10 & 30 & 2 & 20 & - & $10^2$ & $10^{-4}$ & $10^{-2}$ & Y & 0.002 \\
 &  50 & 26 & 3 & 40 & - & $10^3$ & $10^{-5}$ & $10^{-3}$ & N & 0.02 \\ \midrule
\multirow{3}{*}{CIFAR-100} & 1 & 20 & 3 & 20 & - & $10^3$ & $10^{-5}$ & $10^{-2}$ & Y & 0.001 \\
 &  10 & 13 & 3 & 30 & - & $10^3$ & $10^{-5}$ & $10^{-2}$ & N & 0.002 \\
 &  50 & 50 & 2 & 40 & 100 & $10^3$ & $10^{-5}$ & $10^{-2}$ & Y & 0.0002 \\ \bottomrule
\end{tabular}
}
\end{table*}

\begin{table*}[htbp]
\centering
\caption{Optimal hyperparameters for FTD. A synthesis batch size of ‘-’ means that we used the full support set at each synthesis step. }
\label{tab:hyperparametersFTD}

\resizebox{1.0\textwidth}{!}{
\begin{tabular}{c|c|c|c|c|c|c|c|c|c|c|c}
\toprule
Dataset & IPC & \begin{tabular}[c]{@{}c@{}}Synthetic \\ Step\end{tabular} & \begin{tabular}[c]{@{}c@{}}Expert \\ Epoch\end{tabular} & \begin{tabular}[c]{@{}c@{}}Max Start \\ Epoch\end{tabular} & \begin{tabular}[c]{@{}c@{}}Synthetic \\ Batch \\ Size\end{tabular} & \begin{tabular}[c]{@{}c@{}}Learning \\ Rate \\ (Pixels)\end{tabular} & \begin{tabular}[c]{@{}c@{}}Learning \\ Rate \\ (Step Size)\end{tabular} & \begin{tabular}[c]{@{}c@{}}Learning \\ Rate \\ (Teacher)\end{tabular} & \begin{tabular}[c]{@{}c@{}}Balance \\coefficient\end{tabular} & \begin{tabular}[c]{@{}c@{}}EMA \\ Decay\end{tabular} & $\lambda$ \\ \midrule
\multirow{3}{*}{CIFAR-10} & 1 & 50 & 2 & 2 & - & 100 & $10^{-7}$ & 0.01 & 0.3 & 0.9999 & 0.002 \\
 & 10 & 30 & 2 & 20 & - & 100 & $10^{-5}$ & 0.001 & 0.3 & 0.9995 & 0.002 \\
 & 50 & 30 & 2 & 40 & - & 1000 & $10^{-5}$ & 0.001 & 1 & 0.999 & 0.0002 \\ \midrule
\multirow{3}{*}{CIFAR-100} & 1 & 40 & 3 & 20 & - & 1000 & $10^{-5}$ & 0.01 & 1 & 0.9995 & 0.002 \\
 & 10 & 20 & 2 & 40 & - & 1000 & $10^{-5}$ & 0.01 & 1 & 0.9995 & 0.0002 \\
 & 50 & 80 & 2 & 40 & 1000 & 1000 & $10^{-5}$ & 0.01 & 1 & 0.999 & 0.002 \\ \bottomrule
\end{tabular}
}
\end{table*}

\begin{table*}[htbp]
\centering
\caption{Optimal hyperparameters for DATM.}
\label{tab:hyperparametersDATM}
% \vspace{-5pt}

\resizebox{1.0\textwidth}{!}{
\begin{tabular}{c|c|c|c|c|c|c|c|c|c|c|c|c|c}
\toprule
Dataset & Model & IPC & \begin{tabular}[c]{@{}c@{}}Synthetic \\ Step\end{tabular} & \begin{tabular}[c]{@{}c@{}}Expert \\ Epoch\end{tabular} & \begin{tabular}[c]{@{}c@{}}Min Start \\ Epoch\end{tabular} & \begin{tabular}[c]{@{}c@{}}Current \\ Max Start \\ Epoch\end{tabular} & \begin{tabular}[c]{@{}c@{}}Max Start \\ Epoch\end{tabular} & \begin{tabular}[c]{@{}c@{}}Synthetic \\ Batch Size\end{tabular} & \begin{tabular}[c]{@{}c@{}}Learning \\ Rate \\ (Label)\end{tabular} & \begin{tabular}[c]{@{}c@{}}Learning \\ Rate \\ (Pixels)\end{tabular} & \begin{tabular}[c]{@{}c@{}}Num \\ Eval\end{tabular} & \begin{tabular}[c]{@{}c@{}}Eval \\ Iteration\end{tabular} & $\lambda$ \\ \midrule
\multirow{3}{*}{CIFAR-10} & \multirow{3}{*}{ConvNetD3} & 1 & 80 & 2 & 0 & 4 & 4 & 10 & 5 & 100 & 5 & 500 & 0.0002 \\
 &  & 10 & 80 & 2 & 0 & 10 & 20 & 100 & 2 & 100 & 5 & 500 & 0.0005 \\
 &  & 50 & 80 & 2 & 0 & 20 & 40 & 500 & 2 & 1000 & 5 & 500 & 0.002 \\ \midrule
\multirow{1}{*}{CIFAR-100} & \multirow{1}{*}{ConvNetD3} & 1 & 40 & 3 & 0 & 10 & 20 & 100 & 10 & 1000 & 5 & 500 & 0.02 \\ \midrule
\multirow{3}{*}{Tiny ImageNet} & \multirow{3}{*}{ConvNetD4} & 1 & 60 & 2 & 0 & 15 & 20 & 200 & 10 & 10000 & 5 & 500 & 0.002 \\
 &  & 10 & 60 & 2 & 10 & 50 & 50 & 250 & 10 & 100 & 3 & 500 & 0.002 \\
 &  & 50 & 80 & 2 & 40 & 70 & 70 & 250 & 10 & 100 & 3 & 500 & 0.002 \\ \bottomrule
\end{tabular}
}
\end{table*}

\subsection{Limitation}
\label{supp:limitation}
\textbf{Computational Cost}: Similar to other methods, we have not yet addressed the large computational cost associated with the dataset distillation. Our experiments were conducted on a mix of RTX 2080 Ti, RTX 3090, RTX 4090, NVIDIA A100, and NVIDIA V100 GPUs. The cost in terms of computational resources and time remains significant for large datasets and high IPC experiments. For example, distilling Tiny ImageNet using DATM with IPC = 1 requires approximately 150GB of GPU memory, and for IPC = 50, a single experiment can take nearly 24 hours to complete.

\textbf{Hyperparameter Tuning}: The selection of the $\lambda$ requires manual adjustment, which may involve additional costs. The extensive training durations and substantial GPU memory requirements make it challenging to conduct exhaustive experiments with multiple $\lambda$ values to identify the global optimum, given our computational resource limitations. By exploring a wider range of $\lambda$ values, it is possible to obtain better results.

% \textbf{Computational Resources}: Our experiments were conducted on a mix of RTX 2080 Ti, RTX 3090, RTX 4090, NVIDIA A100, and NVIDIA V100 GPUs. Due to our relatively limited computational resources, the pressure of training times, and the heavy reliance on GPU memory, we were unable to thoroughly experiment with multiple $\lambda$ values to find the optimal one. By exploring a wider range of $\lambda$ values, it is possible to attain higher outcomes.

\subsection{Pseudocodes of adding SDC on Matching-based Distillation Methods}\label{supp:algs}
We provide detailed pseudocodes for GM-based methods and TM-based methods. We take DC as the standard GM-based method, and MTT as the standard TM-based method. The detailed pseudocodes are shown in Algorithm~\ref{alg:GM} for GM-based methods and Algorithm~\ref{alg:TM} for TM-based methods.

\begin{algorithm}[h]
    \small
    \caption{Gradient Matching with Sample Difficulty Correction}
    \label{alg:GM}
    \begin{algorithmic}[1]
        \Require Training set $\mathcal{D}_\mathsf{real}$, category set $C$,  classification cross-entropy loss $\mathcal{L}$, probability distribution for weights $P_\theta$, distance metric $\mathbf{D}$, regularization coefficient $\lambda$, number of steps $T$, learning rate $\eta$ for network parameters.
        \State {Initialize distilled data $\mathcal{D}_\mathsf{syn} \sim \mathcal{D}_\mathsf{real}$}.
        \For {\textbf{each} distillation step...}
            \State {$\triangleright$ Initialize network $\theta_0 \sim P_\theta$}
            \For {$t = 0 \to T$}
                \For {$c = 0 \to C-1$}
                    \State $\triangleright$ Sample a mini-batch of distilled images: $\mathcal{B}_{\mathsf{real}}^c \sim \mathcal{D}^c_\mathsf{real}$
                    \State $\triangleright$ Sample a mini-batch of original images: $\mathcal{B}_{\mathsf{syn}}^c \sim \mathcal{D}^c_\mathsf{syn}$
                    \State $\triangleright$ Compute $\mathcal{L}_{\mathcal{B}_{\mathsf{syn}}^c}= \mathbb{E}_{(x,y)\in \tilde{\mathcal{B}}_{\mathsf{real}}^c} \left[\mathcal{L}(x,y;\theta_t)\right]$, $\mathcal{L}_{\mathcal{B}_{\mathsf{real}}^c}= \mathbb{E}_{(x,y)\in \tilde{\mathcal{B}}_{\mathsf{real}}^c} \left[\mathcal{L}(x,y;\theta_t)\right]$
                    \State $\triangleright$ Compute gradient matching loss $L=\mathbf{D}\left(\nabla_\theta \mathcal{L}_{\mathcal{B}_{\mathsf{real}}^c}, \nabla_\theta \mathcal{L}_{\mathcal{B}_{\mathsf{syn}}^c}\right) + \lambda \left\|\nabla_\theta \mathcal{L}_{\mathcal{B}_{\mathsf{syn}}}\right\|_2^2$
                    \State {$\triangleright$ Update $\mathcal{D}_\mathsf{syn}$ \emph{w.r.t.} $L$}
                \EndFor
                \State $\triangleright$ Update network \emph{w.r.t.} classification loss: $\theta_{t+1} = \theta_{t} - \eta\nabla \mathcal{L}_{\mathcal{D}_{\mathsf{syn}}}(\theta_{t})$
            \EndFor
        \EndFor
        \Ensure {distilled data $\mathcal{D}_\mathsf{syn}$}
    \end{algorithmic}
\end{algorithm}

\begin{algorithm}[h]
    \small
    \caption{Trajectory Matching with Sample Difficulty Correction}
    \label{alg:TM}
    \begin{algorithmic}[1]
        \Require Set of expert parameter trajectories trained on $\mathcal{D}_\mathsf{real}$ $\{\tau_i^*\}$, the number of updates between starting and target expert params $M$, the number of updates to student network per distillation step $N$, differentiable augmentation function $\mathcal{A}$, maximum start epoch $T^+ < T$, learning rate $\eta$ for network parameters, regularization coefficient $\lambda$, classification cross-entropy loss $\mathcal{L}$.
        \State {Initialize distilled data $\mathcal{D}_\mathsf{syn} \sim \mathcal{D}_\mathsf{real}$}.
        \For {\textbf{each} distillation step...}
            \State {$\triangleright$ Sample expert trajectory: $\tau^* \sim \{\tau^*_i\}$ with $ \tau^* = \{\theta_{t}^{\mathcal{D}_{\mathsf{real}}}\}_0^T$}
            \State {$\triangleright$ Choose random start epoch, $t \leq T^+$}
            \State {$\triangleright$ Initialize student network with expert params: $\theta_{t}^{\mathcal{D}_{\mathsf{syn}}} := \theta_{t}^{\mathcal{D}_{\mathsf{real}}}$}
            \For {$n = 0 \to N-1$}
                \State $\triangleright$ Sample a mini-batch of distilled images: $\mathcal{B}_{\mathsf{syn}} \sim \mathcal{D}_\mathsf{syn}$
                \State $\triangleright$ Update student network \emph{w.r.t.} classification loss: $\theta_{t+n+1}^{\mathcal{D}_{\mathsf{syn}}} = \theta_{t+n}^{\mathcal{D}_{\mathsf{syn}}} - \eta\nabla \mathcal{L}_{\mathcal{A}(\mathcal{B}_{\mathsf{syn}})}( \theta_{t+n}^{\mathcal{D}_{\mathsf{syn}}})$
            \EndFor
            \State {$\triangleright$ Compute loss between ending student and expert params:}
            $L = \frac{\|\theta_{t+N}^{\mathcal{D}_{\mathsf{syn}}}-\theta_{t+M}^{\mathcal{D}_{\mathsf{real}}}\|_2^2}{\|\theta_{t+M}^{\mathcal{D}_{\mathsf{real}}}- \theta_t^{\mathcal{D}_{\mathsf{real}}}\|_2^2} + \lambda \left\|\nabla_\theta \mathcal{L}_{\mathcal{D}_{\mathsf{syn}}}\right\|_2^2$
            \State {$\triangleright$ Update $\mathcal{D}_\mathsf{syn}$ \emph{w.r.t.} $L$}
        \EndFor
        \Ensure {distilled data $\mathcal{D}_\mathsf{syn}$}
    \end{algorithmic}
\end{algorithm}

\section{Exploring the Effectiveness of SDC in Additional Experiments}
\subsection{More Results on the Cross-architecture Evaluation}\label{supp:cross}

To evaluate the performance of distilled datasets on different network architectures using SDC (marked as +SDC in the tables) and other methods (DATM and DSAC), we conducted cross-architecture evaluation experiments. We compared the effects of DATM and SDC on CIFAR-10, CIFAR-100, and Tiny ImageNet datasets, and the effects of DSAC and SDC on MNIST, FashionMNIST, and SVHN datasets. Finally, we further evaluated the performance differences between DSAC and SDC methods on MNIST, FashionMNIST, and SVHN datasets with IPC = 50. The cross-architecture evaluation experiments for DSAC and DATM, as well as the use of the SDC method on datasets with IPC = 1 of DATM and IPC = 10 of DSAC, can be found in Table~\ref{table:cross}.

The results of evaluating distilled datasets learned through DATM and SDC methods on CIFAR-10, CIFAR-100, and Tiny ImageNet datasets using ResNet-18, VGG-11, AlexNet, LeNet, and MLP networks are presented in Table~\ref{table:tm1}. For instance, on the CIFAR-100 dataset, the accuracy of the VGG-11 network improved by \textbf{1.46\%}. It can be observed that the performance after applying SDC is generally better than DATM.

In our evaluation of distilled datasets learned through DSAC and SDC methods on MNIST, FashionMNIST, and SVHN datasets using the same network architectures, as detailed in Table~\ref{table:gm1}, the results show that performance after applying SDC is superior to the DSAC method across datasets and network architectures. For example, on the MNIST dataset, the accuracy of the VGG-11 network improved by \textbf{1.37\%}, and on the SVHN dataset, the accuracy of the ResNet-18 improved by \textbf{0.61\%}.

Additionally, similar evaluation results on MNIST, FashionMNIST, and SVHN datasets with an IPC value of 50 are summarized in Table~\ref{table:gm50}. For example, on the SVHN dataset, the accuracy of the LeNet network improved by \textbf{1.0\%}. It can be seen that with an increase in IPC value, the performance after applying SDC remains better in most cases, further demonstrating the superiority of the SDC method in dataset distillation.

\begin{table*}[htbp]
  \caption{Cross-architecture evaluation. We evaluated distilled datasets with IPC = 10 learned through  DATM w/ and w/o {\mymethod} on different networks.}
  \vspace{5pt}
  \label{table:tm1}
  \centering

  \resizebox{0.8\textwidth}{!}{
\begin{tabular}{cc|ccccc}
    \toprule
     Dataset & Method & ResNet-18 & VGG-11 & AlexNet & LeNet & MLP \\
     \midrule
      \multirow{2}{*}{CIFAR-10} & DATM & 36.48& 37.32& 33.19& 32.56& 27.21\\
      & \textbf{{+\mymethod}} &\textbf{38.33}& \textbf{38.22}& \textbf{34.56}& \textbf{33.17}& \textbf{27.62}\\
     % \midrule
      \multirow{2}{*}{CIFAR-100} & DATM & 17.87& 14.71& 15.09& 11.76& 11.52\\
      & \textbf{{+\mymethod}} &\textbf{18.97}& \textbf{16.17}& \textbf{15.73}& \textbf{12.44}& \textbf{11.87}\\
     % \midrule
      \multirow{2}{*}{Tiny ImageNet} & DATM & 6.33& 8.67& 6.18& 3.65& 3.34\\
      & \textbf{{+\mymethod}} &\textbf{7.20}&\textbf{9.13}& \textbf{6.89}& \textbf{3.88}& \textbf{3.40}\\
    \bottomrule
  \end{tabular}

}

\vspace{15pt}
\end{table*}

% \begin{table*}[htbp]
%   \caption{Cross-architecture evaluation results of DATM and the application of {\mymethod} on different network architectures (IPC = 50).Cross-architecture evaluation for TM-based methods..}
%   \vspace{-5pt}
%   \label{table:cross-tm}
%   \centering

%   \resizebox{0.8\textwidth}{!}{
%   \begin{tabular}{cc|ccccc}
%     \toprule
%      Dataset & Method & ResNet-18 & VGG-11 & AlexNet & LeNet & MLP \\
%      \midrule
%       \multirow{2}{*}{CIFAR-10} & DATM & \textbf{31.76} & 49.41 & 40.99 & 34.60 & \textbf{31.76} \\
%       & +SDC & 30.87 & \textbf{52.00} & \textbf{41.30} & \textbf{35.40} & 30.87 \\
%       \midrule
%       \multirow{2}{*}{CIFAR-100} & DATM & 25.81 & \textbf{21.59} & \textbf{21.70} & 18.96 & 17.61 \\
%       & +SDC & \textbf{25.91} & 21.06 & 21.56 & \textbf{19.15} & \textbf{18.06} \\
%       \midrule
%       \multirow{2}{*}{Tiny ImageNet} & DATM & xxx & xxx & xxx & xxx & xxx \\
%       & +SDC & \textbf{xxx} & \textbf{xxx} & \textbf{xxx} & \textbf{xxx} & \textbf{xxx} \\
%     \bottomrule
%   \end{tabular}
% }

% \vspace{-15pt}
% \end{table*}

\begin{table*}[htbp]
  \caption{Cross-architecture evaluation. We evaluated distilled datasets with IPC = 1 learned through  DSAC w/ and w/o {\mymethod} on different networks.}
  \label{table:gm1}
  \centering

  \resizebox{0.8\textwidth}{!}{
  \begin{tabular}{cc|ccccc}
    \toprule
     Dataset & Method & ResNet-18 & VGG-11 & AlexNet & LeNet & MLP \\
     \midrule
      \multirow{2}{*}{MNIST} & DSAC & 88.58 & 79.58 & 83.63 & 83.46 & 72.78 \\
      & \textbf{+SDC} & \textbf{88.70} & \textbf{80.95} & \textbf{83.92} & \textbf{83.66} & \textbf{73.51} \\
      \midrule
      \multirow{2}{*}{FashionMNIST} & DSAC & 71.60 & 68.03 & 66.03 & 67.09 & 63.85 \\
      & \textbf{+SDC} & \textbf{71.70} & \textbf{68.82} & \textbf{66.47} & \textbf{67.19} & \textbf{64.93} \\
      \midrule
      \multirow{2}{*}{SVHN} & DSAC & 33.04 & 32.32 & 14.63 & 20.89 & 13.32 \\
      & \textbf{+SDC} & \textbf{33.65} & \textbf{33.84} & \textbf{17.18} & \textbf{22.40} & \textbf{13.86} \\
    \bottomrule
  \end{tabular}
}

\end{table*}

\begin{table*}[htbp]
  \caption{Cross-architecture evaluation. We evaluated distilled datasets with IPC = 50 learned through  DSAC w/ and w/o {\mymethod} on different networks.}
  \label{table:gm50}
  \centering

  \resizebox{0.8\textwidth}{!}{
  \begin{tabular}{cc|ccccc}
    \toprule
     Dataset & Method & ResNet-18 & VGG-11 & AlexNet & LeNet & MLP \\
     \midrule
      \multirow{2}{*}{MNIST} & DSAC & \textbf{97.97} & 98.53 & 97.95 & 97.58 & 94.70 \\
      & \textbf{+SDC} & 97.95 & \textbf{98.57} & \textbf{97.97} & \textbf{97.62} & \textbf{94.75} \\
      \midrule
      \multirow{2}{*}{FashionMNIST} & DSAC & 86.92 & 87.03 & \textbf{85.61} & 84.96 & 83.56 \\
      & \textbf{+SDC} & \textbf{86.96} & \textbf{87.20} & 85.58 & \textbf{85.36} & \textbf{83.78} \\
      \midrule
      \multirow{2}{*}{SVHN} & DSAC & 86.10 & 85.62 & 83.47 & 77.92 & 62.68 \\
      & \textbf{+SDC} & \textbf{86.32} & \textbf{85.86} & \textbf{83.85} & \textbf{78.92} & \textbf{63.47} \\
    \bottomrule
  \end{tabular}
}

\end{table*}

\subsection{More Results on the Adaptive Sample Difficulty Correction}\label{supp:asdc}

\begin{figure}[htbp]
    \centering
    \includegraphics[width=0.45\textwidth]{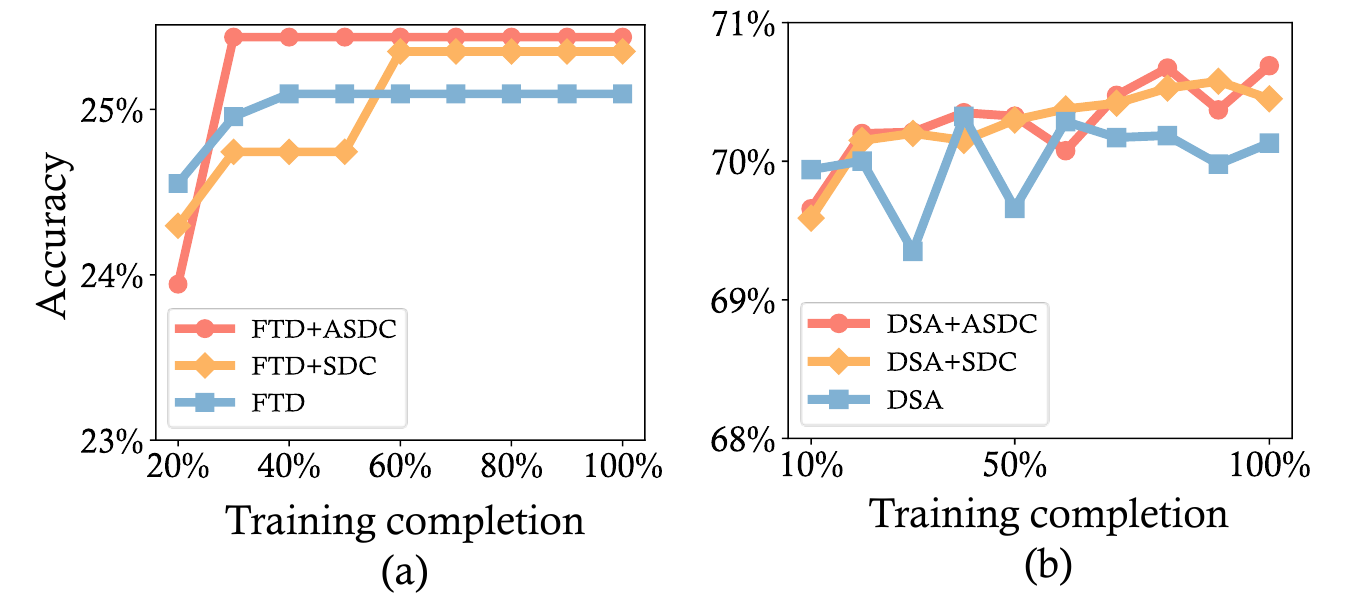}
    \caption{The application of ASDC on (a) FTD  and (b) DSA. Gradually increasing $\lambda$ gives better results than the baseline and the method after applying SDC.}
    \label{fig:increasing_lambda_supp}
\end{figure}
The dynamic adjustment of {\mymethod}, when applied to both DSA and FTD, consistently outperforms both the baseline methods and the baseline methods with {\mymethod} applied. As shown in Figure~\ref{fig:increasing_lambda_supp}, we logarithmically increased the $\lambda$ coefficient for DSA from 0.0002 to 0.002 over 1000 steps and for FTD from 0.002 to 0.008 over 10,000 iterations. The results clearly demonstrate that ASDC yields superior performance.
Flexibly adjusting the sample difficulty correction by adaptively increasing $\lambda$ yields higher accuracy compared to the standard {\mymethod} and baseline methods.

\subsection{Sensitivity Analysis of SDC coefficient .}\label{supp:sensitivity}
\begin{figure}[htbp]
    \centering
    \includegraphics[width=0.45\textwidth]{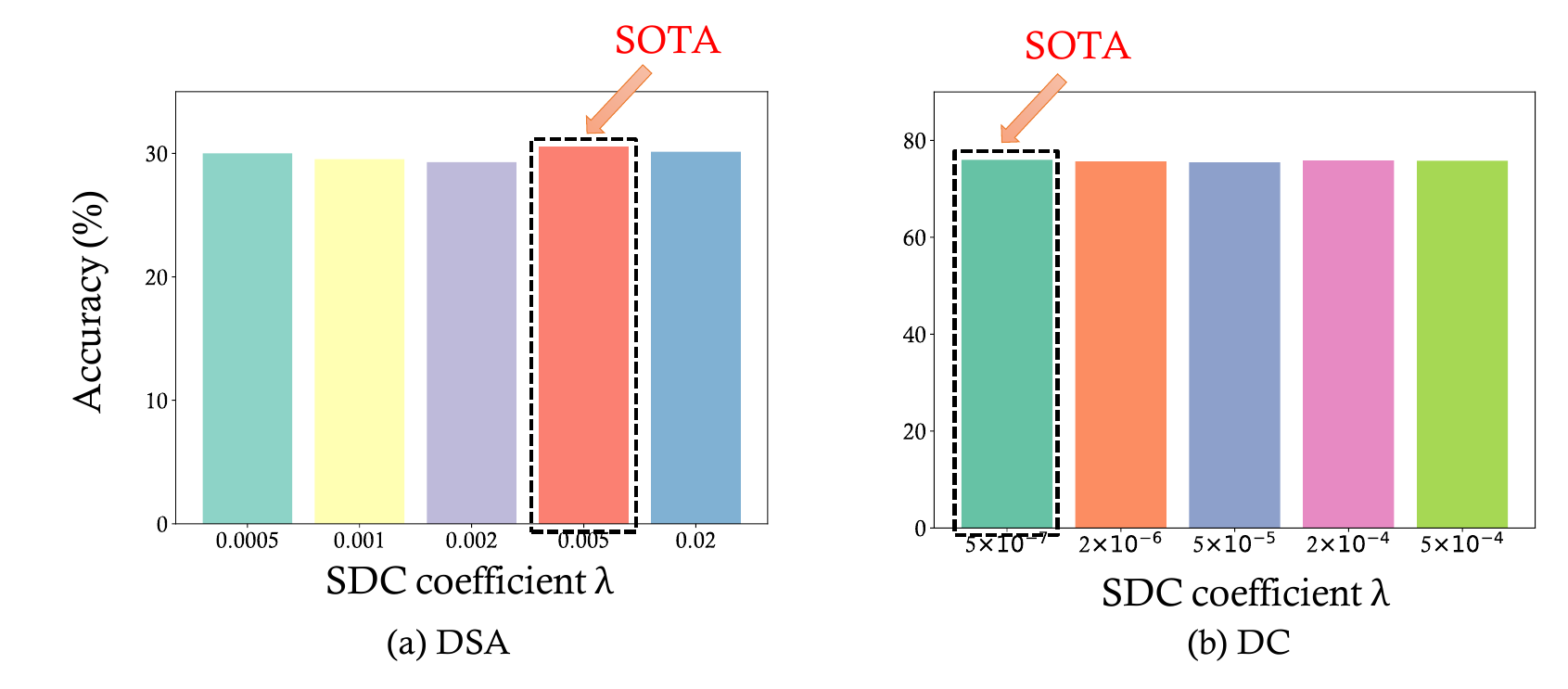}
    \caption{Sensitivity analysis of SDC coefficient $\lambda$ on different distillation methods. We evaluated the sensitivity across different $\lambda$s, and showed that the choice of $\lambda$ did not severely affect the final test performance. (a) DSA on SVHN dataset with IPC = 1 (b) DC on SVHN dataset with IPC = 10.}
    \label{fig:sensitivity}
\end{figure}

In this section, we conducted extensive experiments to study the sensitivity of the hyperparameter $\lambda$. Specifically, we conducted experiments of DSA on SVHN dataset with IPC = 1, and DC on SVHN dataset with IPC = 10. As shown in Figure~\ref{fig:sensitivity}, the choice of $\lambda$ is not sensitive among different matching-based dataset distillation methods.

\section{Analytical Theory for Dataset Distillation}\label{supp:theory}
In this section, we introduce a theory, adapted from data pruning \cite{sorscher2022beyond}, to the context of dataset distillation within an expert-student perceptron framework, utilizing the tools of statistical mechanics. We investigate the challenge of classifying a dataset $\mathcal{D}_{\mathsf{real}}$ consisting of $d_{\mathsf{real}}$ samples $\{x_i, y_i\}_{i=1,\ldots,d_{\mathsf{real}}}$, where the inputs $x_i \sim \mathcal{N}(0, I_{d})$ are i.i.d. zero-mean, unit-variance random Gaussian variables, and the labels $y_i = \text{sign}(\theta^{\mathcal{D}_\mathsf{real}^\top} x_i)$ are generated by an expert perceptron $\theta^{\mathcal{D}_\mathsf{real}} \in \mathbb{R}^{d}$. We assume that the expert perceptron $\theta^{\mathcal{D}_\mathsf{real}}$ is randomly drawn from a uniform distribution on the sphere $\theta^{\mathcal{D}_\mathsf{real}} \sim \text{Unif}(\mathbb{S}^{d-1}(\sqrt{d}))$. Our analysis is situated within the high-dimensional statistics limit where $d, d_{\mathsf{real}} \to \infty$ but the ratio $\alpha_{\mathsf{real}} = d_\mathsf{real} / d$ remains $O(1)$.

Specifically, consider synthesizing a dataset by matching only the samples with the smallest margin $|z_i| = |\theta^{\mathsf{probe}^\top} x_i|$ along a probe student $\theta^{\mathsf{probe}}$. The distilled dataset will then follow a distribution $p(z)$ in the direction of $\theta^{\mathsf{probe}}$ while remaining isotropic in the null space of $\theta^{\mathsf{probe}}$. We assume, without loss of generality, that $\theta^{\mathsf{probe}}$ has developed some overlap with the expert, quantified by the angle $\gamma = \cos^{-1}\left(\frac{\theta^{\mathsf{probe}^\top} \theta^{\mathcal{D}_\mathsf{real}}}{\|\theta^{\mathsf{probe}}\|_2 \|\theta^{\mathcal{D}_\mathsf{real}}\|_2}\right)$.

Once the dataset has been distilled, we consider training a new student $\theta^{\mathcal{D}_\mathsf{syn}}$ from scratch on this distilled dataset. A typical training algorithm aims to find the solution $\theta^{\mathcal{D}_\mathsf{syn}}$ which classifies the training data with maximal margin $\kappa = \min_{i}(\theta^{\mathcal{D}_\mathsf{syn}^\top} y_i x_i)$. Our goal is to compute the generalization error $\varepsilon$ of this student, governed by the overlap between the student and the expert: $\varepsilon = \cos^{-1}(R) / \pi$, where $R = \frac{\theta^{\mathcal{D}_\mathsf{syn}^\top} \theta^{\mathcal{D}_\mathsf{real}}}{\|\theta^{\mathcal{D}_\mathsf{syn}}\|_2 \|\theta^{\mathcal{D}_\mathsf{real}}\|_2}$.

We provide saddle point equations for the cosine similarity $R$ between the probe $\theta^{\mathcal{D}_\mathsf{probe}}$ and the expert $\theta^{\mathcal{D}_\mathsf{real}}$, which will be discussed in Section~\ref{supp:perfect} and Section~\ref{supp:imperfect}. For our simulations, we set the parameter dimension $d = 200$ for perfect probe settings, and set $d=50$ for imperfect probe settings. We averaged 100 simulation results to verify the theory.

\subsection{Perfect Expert-Teacher Settings}\label{supp:perfect}

The solution is given by the following saddle point equations for perfect expert-teacher settings, \emph{i.e.,} $\gamma=0$. For any given $ \alpha_{\mathsf{syn}}$, these equations can be solved for the order parameters $ R, \kappa $. From these parameters, the generalization error can be computed as $ \varepsilon = \cos^{-1}(R)/\pi $. 

\begin{align*}
R &= \frac{2\alpha_{\mathsf{syn}}}{f\sqrt{2\pi}\sqrt{1-R^{2}}} \int_{-\infty}^{\kappa} Dt \ \exp\left(-\frac{R^{2}t^{2}}{2(1-R^{2})}\right) \\
&\quad \times \left[1-\exp\left(-\frac{\gamma(\gamma-2Rt)}{2(1-R^{2})}\right)\right] (\kappa-t) 
\end{align*}

\begin{align*}
1-R^{2} &= \frac{2\alpha_{\mathsf{syn}}}{f} \int_{-\infty}^{\kappa} Dt \ \bigg[H\left(-\frac{Rt}{\sqrt{1-R^{2}}}\right) \\
&\quad - H\left(-\frac{Rt-\gamma}{\sqrt{1-R^{2}}}\right)\bigg] (\kappa-t)^{2}
\end{align*}

Where $H(x) = \frac{1}{2} \left( 1 - \frac{2}{\sqrt{\pi}} \int_{0}^{\left(\frac{x}{\sqrt{2}}\right)} e^{-t^2} \, dt  \right).$ 
This calculation produces the solid theoretical curves shown in Figure~\ref{fig:theory}, which exhibit an excellent match with numerical simulations.
Please refer \cite{sorscher2022beyond} for detailed deductions.

\subsection{Imperfect Expert-Teacher Settings}\label{supp:imperfect}

We have shown the perfect student settings in Section~\ref{supp:perfect}. When the probe student does not exactly match the expert, an additional parameter $\theta$ characterizes the angle between the probe student and the expert. Furthermore, an additional order parameter $\rho = \theta^{\mathcal{D}_\mathsf{real}^\top} \theta^{\mathcal{D}_\mathsf{syn}}$ represents the typical student-probe overlap, which must be optimized. Consequently, we derive three saddle point equations.
\begin{scriptsize}
\begin{align*}
\frac{R-\rho\cos\gamma}{\sin^{2}\gamma} &= \frac{\alpha_{\mathsf{syn}}}{\pi\Lambda} \left\langle \int_{-\infty}^{\kappa} dt \ \exp\left(-\frac{\Delta(t,z)}{2\Lambda^{2}}\right) \right. \\
&\quad \left. \times (\kappa-t) \right\rangle_{z} \\
1-\frac{\rho^{2}+R^{2}-2\rho R\cos\gamma}{\sin^{2}\gamma} &= 2\alpha_{\mathsf{syn}} \left\langle \int_{-\infty}^{\kappa} dt \ \frac{e^{-\frac{(t-\rho z)^{2}}{2(1-\rho^{2})}}}{\sqrt{2\pi}\sqrt{1-\rho^{2}}} \right. \\
&\quad \left. \times H\left(\frac{\Gamma(t,z)}{\sqrt{1-\rho^{2}}\Lambda}\right) (\kappa-t)^{2} \right\rangle_{z} \\
\frac{\rho-R\cos\gamma}{\sin^{2}\gamma} &= 2\alpha_{\mathsf{syn}} \left\langle \int_{-\infty}^{\kappa} dt \ \frac{e^{-\frac{(t-\rho z)^{2}}{2(1-\rho^{2})}}}{\sqrt{2\pi}\sqrt{1-\rho^{2}}} \right. \\
&\quad \left. \times H\left(\frac{\Gamma(t,z)}{\sqrt{1-\rho^{2}}\Lambda}\right) \left(\frac{z-\rho t}{1-\rho^{2}}\right) (\kappa-t) \right\rangle_{z} \\
&\quad + \frac{1}{2\pi\Lambda} \left\langle \exp\left(-\frac{\Delta(t,z)}{2\Lambda^{2}}\right) \right. \\
&\quad \left. \times \left(\frac{\rho R-\cos\gamma}{1-\rho^{2}}\right) (\kappa-t) \right\rangle_{z}
\end{align*}
\end{scriptsize}
Where,
\begin{align*}
\Lambda &= \sqrt{\sin^{2}\gamma-R^{2}-\rho^{2}+2\rho R\cos\gamma}, \\
\Gamma(t,z) &= z(\rho R-\cos\gamma)-t(R-\rho\cos\gamma), \\
\Delta(t,z) &= z^{2}\left(\rho^{2}+\cos^{2}\gamma-2\rho R\cos\gamma\right) \\
&\quad + 2tz(R\cos\gamma-\rho) + t^{2}\sin^{2}\gamma.
\end{align*}

\begin{figure*}[tb!]
    \centering
    \includegraphics[width=0.99\textwidth]{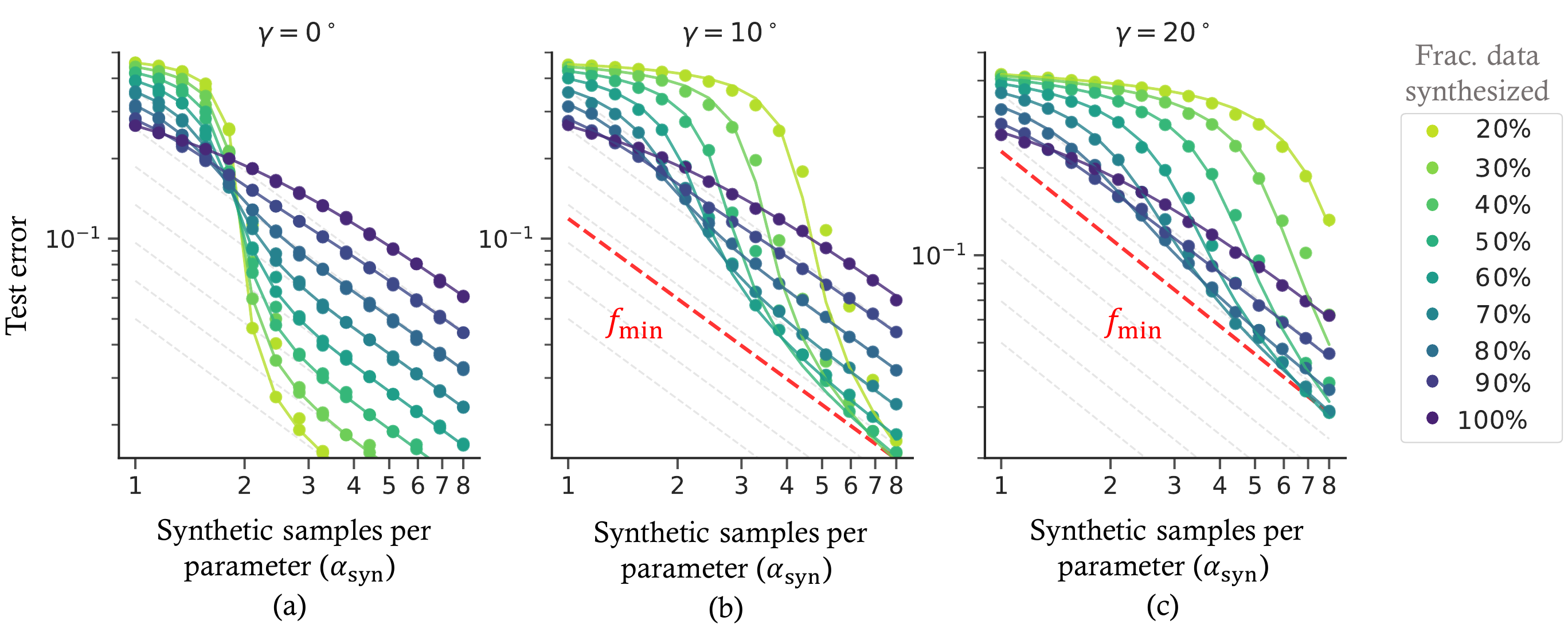}
    \caption{Test error $\varepsilon$ as a function of the synthetic samples per parameter $\alpha_{\mathsf{syn}}$ and fraction of data synthesized $f$  in (a) the perfect expert setting ($\gamma=0$) (b) the perfect expert setting ($\gamma=10^\circ$) (c) the perfect expert setting ($\gamma=20^\circ$).}
    \label{fig_supp:imperfect}
\end{figure*}

The notation $\langle \cdot \rangle_z$ denotes an average over the pruned data distribution $ p(z) $ for the probe student. For any given $ \alpha_{\mathsf{syn}}, p(z), \gamma $, these equations can be solved for the order parameters $ R, \rho, \kappa $. From these parameters, the generalization error can be readily obtained as $ \varepsilon = \cos^{-1}(R)/\pi $. Our simulation results are shown in Figure~\ref{fig_supp:imperfect}. Please refer \cite{sorscher2022beyond} for detailed deductions.

\section{Visualization Results}\label{supp:visualization}
\begin{figure*}[htbp]
    \centering
    \includegraphics[width=0.7\textwidth]{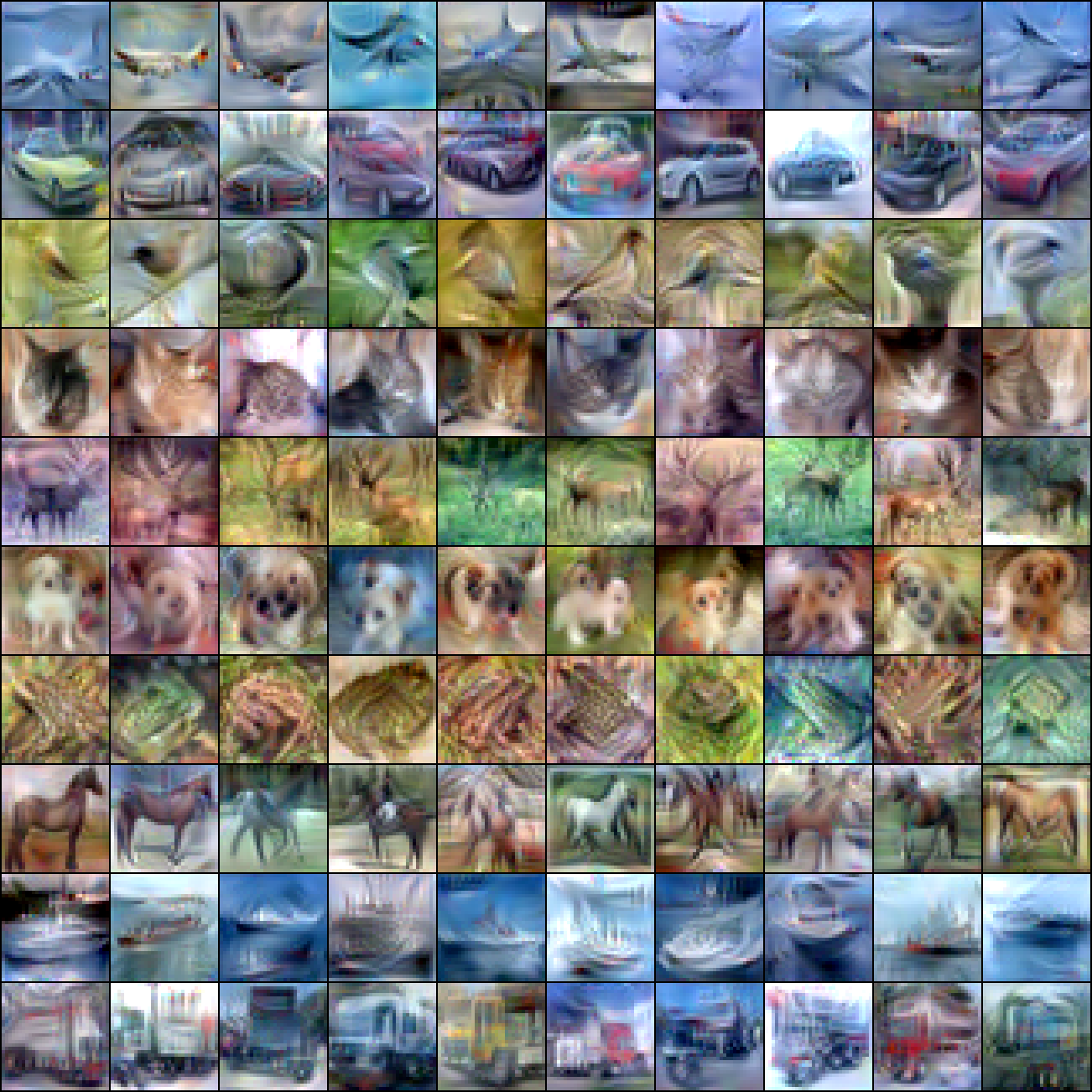}
    \caption{(FTD + SDC, CIFAR-10, IPC = 10) Visualization of distilled images.}
    \label{fig:ftd_c10_ipc10_ours}
\end{figure*}

\begin{figure*}[htbp]
    \centering
    \includegraphics[width=0.7\textwidth]{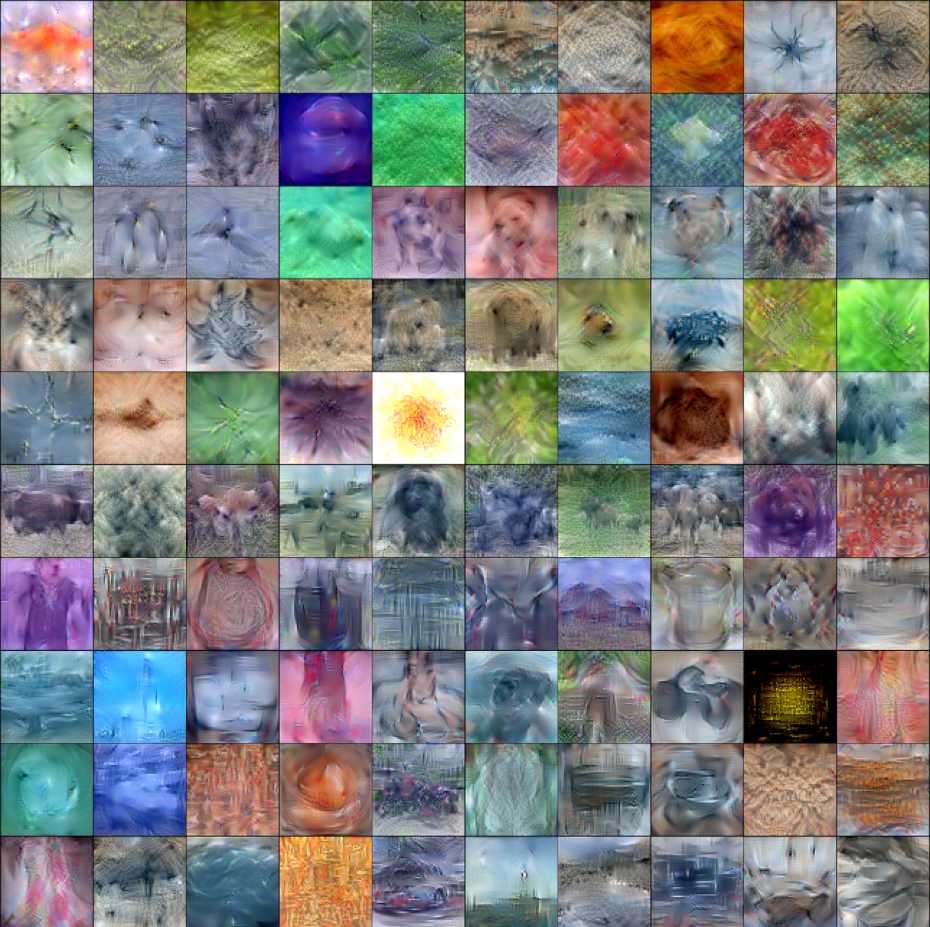}
    \caption{(DATM + SDC, Tiny ImageNet, IPC = 1, 1 / 2) Visualization of distilled images.}
    \label{fig:datm_tiny_ipc1_ours_11}
\end{figure*}

\begin{figure*}[htbp]
    \centering
    \includegraphics[width=0.7\textwidth]{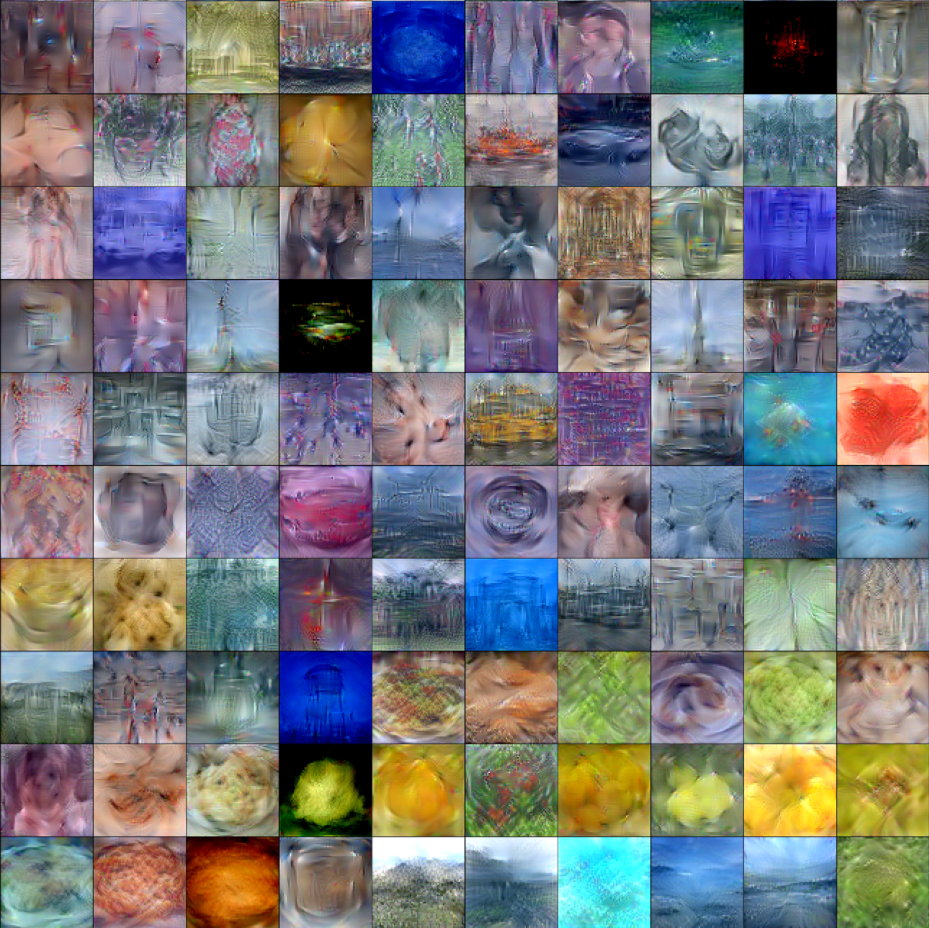}
    \caption{(DATM + SDC, Tiny ImageNet, IPC = 1, 2 / 2) Visualization of distilled images.}
    \label{fig:datm_tiny_ipc1_ours_22}
\end{figure*}

\begin{figure*}[htbp]
    \centering
    \includegraphics[width=0.7\textwidth]{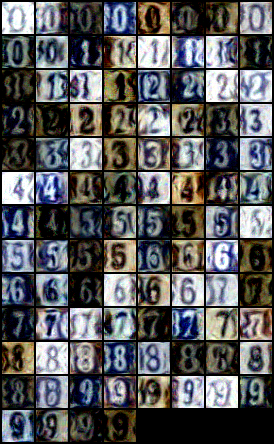}
    \caption{(DSA + SDC, SVHN, IPC = 10) Visualization of distilled images.}
    \label{fig:dsa_svhn_ipc10_ours}
\end{figure*}

\begin{figure*}[htbp]
    \centering
    \includegraphics[width=0.7\textwidth]{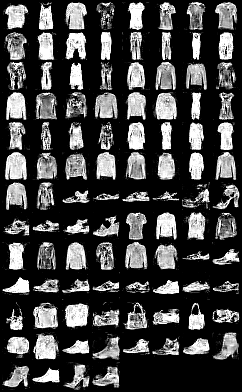}
    \caption{(DC + SDC, FashionMNIST, IPC = 10) Visualization of distilled images.}
    \label{fig:dc_fashionmnist_ipc10_ours}
\end{figure*}

Additionally, we show our visualization of distilled datasets by adding SDC into current matching-based methods, as shown in Figure~\ref{fig:ftd_c10_ipc10_ours}, Figure~\ref{fig:datm_tiny_ipc1_ours_11}, Figure~\ref{fig:datm_tiny_ipc1_ours_22}, Figure~\ref{fig:dsa_svhn_ipc10_ours}, and Figure~\ref{fig:dc_fashionmnist_ipc10_ours}. 
% \begin{figure*}[htbp]
%     \centering
%     \includegraphics[width=0.99\textwidth]{figures/datm_c10_ipc10_ours.png}
%     \caption{(DATM + SDC, CIFAR10, IPC = 10) Visualization of distilled images.}
%     \label{fig:datm_c1o_ipc10_ours}
% \end{figure*}
% \begin{figure*}[htbp]
%     \centering
%     \includegraphics[width=0.99\textwidth]{figures/datm_c10_ipc10_none.png}
%     \caption{(DATM,CIFAR10,IPC = 10)Visualization of distilled images.}
%     \label{fig:datm_c1o_ipc10_none}
% \end{figure*}
% \begin{figure*}[htbp]
%     \centering
%     \includegraphics[width=0.5\textwidth]{figures/datm_tiny_ipc1_ours.png}
%     \caption{(DATM + SDC, Tiny ImageNet, IPC = 1) Visualization of distilled images.}
%     \label{fig:datm_tiny_ipc1_ours}
% \end{figure*}


\clearpage
\newpage

\end{document}